\documentclass[journal]{IEEEtran}
\usepackage[linesnumbered,ruled]{algorithm2e}
\usepackage[usenames, dvipsnames]{color}
\usepackage{graphicx,amssymb,mathrsfs,amsmath,array}
\usepackage{arydshln}
\usepackage{booktabs}
\usepackage{multirow}
\usepackage{color}
\usepackage{xcolor}
\usepackage{graphicx}
\usepackage{colortbl}
\usepackage{subfigure}
\usepackage{amsmath,amssymb} 
\usepackage{makecell}
\usepackage{bbding}
\usepackage{tabularx}
\usepackage{subcaption}
\usepackage[pagebackref=true,breaklinks=true,colorlinks,bookmarks=false]{hyperref}
\usepackage{bm}
\usepackage{url}
\usepackage{tikz}
\usetikzlibrary{spy}
\usepackage{placeins}
\usepackage{utfsym}
\usepackage{bbding}
\usepackage{pifont}
\usepackage{graphics, graphicx}
\usepackage{subfigure}
\usepackage{multirow}
\usepackage{booktabs}
\usepackage{xcolor}
\usepackage{cases}
\usepackage{epstopdf}
\graphicspath{{figures/}}

\usepackage{algpseudocode}
\usepackage{cite}

\begin{document}
\title{UrbanSAM: Learning Invariance-Inspired Adapters for Segment Anything Models in Urban Construction}

\author{Chenyu~Li,
        Danfeng~Hong,~\IEEEmembership{Senior Member,~IEEE,}
        Bing Zhang,~\IEEEmembership{Fellow,~IEEE,}
        Yuxuan~Li,
        Gustau~Camps-Valls,~\IEEEmembership{Fellow,~IEEE,}
        Xiao Xiang Zhu,~\IEEEmembership{Fellow,~IEEE,}
        and Jocelyn~Chanussot,~\IEEEmembership{Fellow,~IEEE}

\thanks{This work was supported by the National Natural Science Foundation of China under Grant 42271350 and the AXA project, which was also supported by the International Partnership Program of the Chinese Academy of Sciences under Grant No.313GJHZ2023066FN.}
\thanks{C. Li is with the School of Mathematics and Statistics, Southeast University, 211189 Nanjing, China. (e-mail: chenyuli.erh@gmail.com)}
\thanks{D. Hong and Y. Li are with the Aerospace Information Research Institute, Chinese Academy of Sciences, Beijing 100094, China, and also with the School of Electronic, Electrical and Communication Engineering, University of Chinese Academy of Sciences, Beijing 100049, China. (e-mail: hongdf@aircas.ac.cn)}
\thanks{B. Zhang is with the Aerospace Information Research Institute, Chinese Academy of Sciences, 100094 Beijing, China, and the College of Resources and Environment, University of Chinese Academy of Sciences, Beijing 100049, China, and also with the School of Mathematics and Statistics, Southeast University, 211189 Nanjing, China. (e-mail: zb@radi.ac.cn)}
\thanks{G. Camps-Valls is with the Image Processing Laboratory (IPL), Universitat de Val\`encia, Paterna, Val\`encia 46980, Spain. (e-mail: gustau.camps@uv.es)}
\thanks{X. X. Zhu is with the Data Science in Earth Observation, Technical University of Munich, Munich 80333, Germany. (e-mail: xiaoxiang.zhu@tum.de)}
\thanks{J. Chanussot is with Univ. Grenoble Alpes, Inria, CNRS, Grenoble INP, LJK, Grenoble 38000, France. (e-mail: jocelyn.chanussot@grenoble-inp.fr)}
}

\maketitle

\begin{abstract}
Object extraction and segmentation from remote sensing (RS) images is a critical yet challenging task in urban environment monitoring. Urban morphology is inherently complex, with irregular objects of diverse shapes and varying scales. These challenges are amplified by heterogeneity and scale disparities across RS data sources, including sensors, platforms, and modalities, making accurate object segmentation particularly demanding. While the Segment Anything Model (SAM) has shown significant potential in segmenting complex scenes, its performance in handling form-varying objects remains limited due to manual-interactive prompting. To this end, we propose UrbanSAM, a customized version of SAM specifically designed to analyze complex urban environments while tackling scaling effects from remotely sensed observations. Inspired by multi-resolution analysis (MRA) theory, UrbanSAM incorporates a novel learnable prompter equipped with a Uscaling-Adapter that adheres to the invariance criterion, enabling the model to capture multiscale contextual information of objects and adapt to arbitrary scale variations with theoretical guarantees. Furthermore, features from the Uscaling-Adapter and the trunk encoder are aligned through a masked cross-attention operation, allowing the trunk encoder to inherit the adapter's multiscale aggregation capability. This synergy enhances the segmentation performance, resulting in more powerful and accurate outputs, supported by the learned adapter. Extensive experimental results demonstrate the flexibility and superior segmentation performance of the proposed UrbanSAM on a global-scale dataset, encompassing scale-varying urban objects such as buildings, roads, and water. The datasets and codes will be openly available at \url{https://github.com/danfenghong} to ensure reproducibility.
\end{abstract}

\graphicspath{{figures/}} 

\begin{figure*}[!t]
      \centering
	   \includegraphics[width=1.0\textwidth]{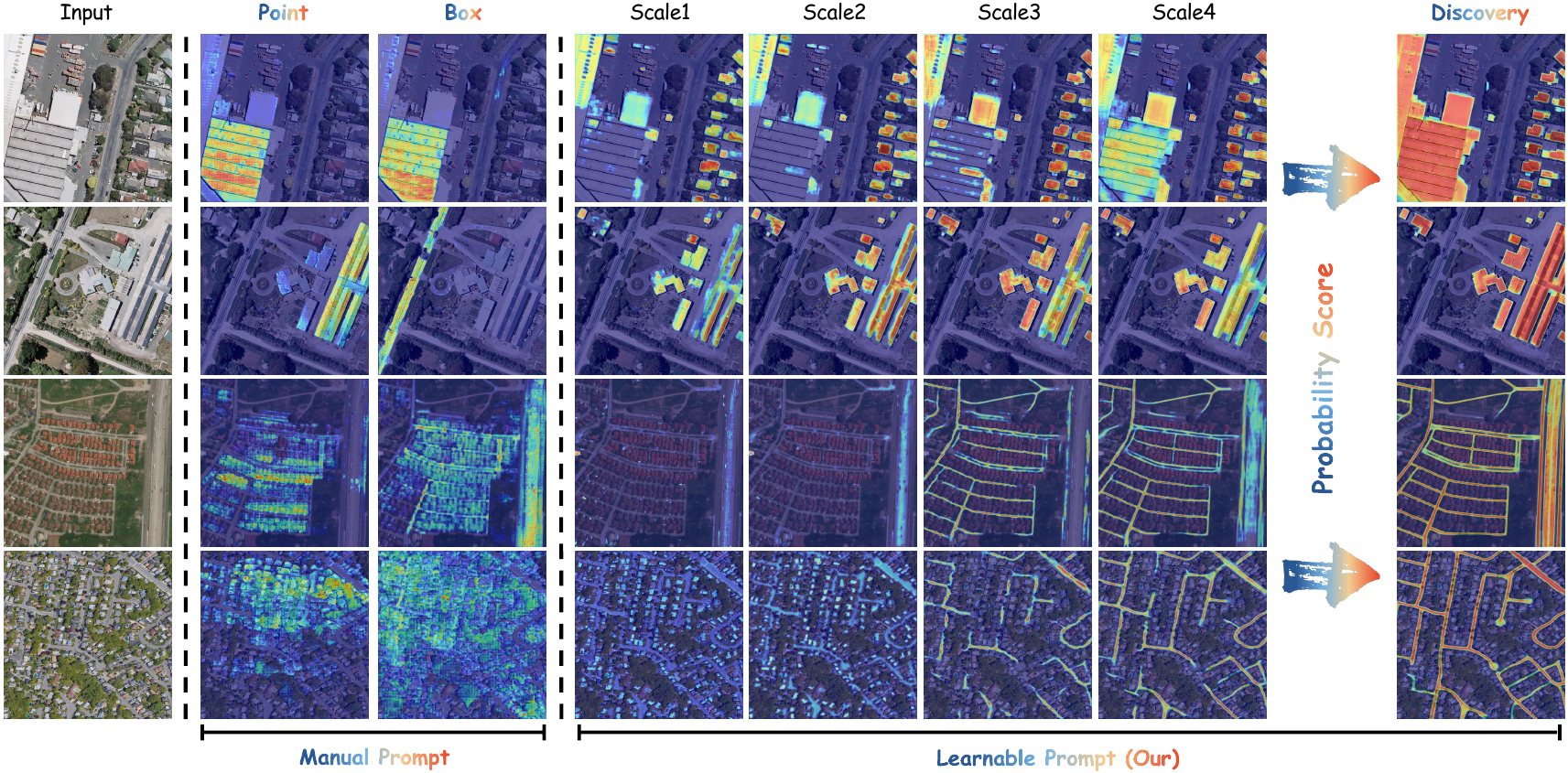}
      \caption{UrbanSAM achieves robust handling of scale variations and superior segmentation performance owing to its invariance-aware adapters inspired by MRA theory. This design effectively addresses scale effects, enhancing adaptability and accuracy in urban segmentation tasks. Notably, manual prompts (point or box) often produce incomplete, low-activation attention maps with high errors and obvious noise, whereas UrbanSAM's learned invariance-aware adapters precisely localize the relevant elements or objects by accumulating and integrating attention across different scales.}
\label{fig:motivation}
\end{figure*} 

\section{Introduction}\label{introduction}
\IEEEPARstart{S}{patial} distributions of urban surface features unveil intricate interactions between natural environments and socio-economic factors, offering critical insights for urban management applications such as urban planning, disaster response, demographics, and transportation. With unparalleled wide-area coverage, RS images emerge as a key data source for efficient large-scale urban analysis \cite{hong2023cross}. Recent advancements in aerospace and sensor technologies have significantly improved the quantity, spatial resolution, and temporal frequency of RS data, unlocking new opportunities for high-precision urban surface mapping. However, this rapid progress has introduced the following challenges that demand further attention:
\begin{itemize}
    \item \textbf{Challenge 1 (\textit{cf.} High Heterogeneity and Complex Object Characteristics):} Urban scenes consist of diverse object types with substantial variations in spectral, textural, and morphological features. This heterogeneity significantly amplifies the complexity of segmentation tasks, as traditional methods struggle to generalize across objects of varying appearances and structures.
    \item \textbf{Challenge 2 (\textit{cf.} High Resolution and Fine-Grained):} Advances in spatial resolution facilitate the finer delineation of urban surface features, enhancing the ability to capture intricate details. However, this improvement also introduces challenges, such as redundant information and increased noise interference, necessitating a careful balance between resolution, computational efficiency, and segmentation accuracy.
    \item \textbf{Challenge 3 (\textit{cf.} High Spatial Density and variegated Layouts):} Cultural and natural differences contribute to the diversity of global urban environments, where internal geographical structures exhibit a wide range of sizes, shapes, and densely packed patterns. This spatial complexity hinders the delineation of distinct object categories, making accurate segmentation particularly challenging.
\end{itemize}

To effectively address the aforementioned challenges in urban surface segmentation, it is essential to develop models that can seamlessly capture the diverse, multiscale, and heterogeneous nature of urban environments. Foundation models \cite{radford2018improving} have sparked a technological revolution in artificial intelligence, leveraging large-scale pre-training on massive datasets to enable generalizable understanding and representation capabilities \cite{kenton2019bert, solaiman2019release, brown2020language, bommasani2021opportunities}. These capabilities can be effectively transferred across diverse domains and downstream applications \cite{dubey2024llama, hong2024spectralgpt}, driving significant advancements in AI research and practice. Propelled by this rapid progress, SAM \cite{kirillov2023segment}, the first foundational model explicitly designed for image segmentation, pre-train on millions of images and over a billion masks. The advent of SAM has ushered in a new paradigm of prompt-driven segmentation, which facilitates its applied in a wide range of expertise, computer vision \cite{wang2023hifi, bhat2024loosecontrol}, medical imaging \cite{hu2024skinsam, wu2023medical, zhang2023customized}, industrial inspection \cite{li2024clipsam, zhang2024gpt, cao2023segment}, and autonomous driving \cite{zhou2023event, luo2023calibanything, li2024fusionsam}. Currently, SAM is being actively explored in the RS domain to enhance segmentation task performance with high-quality predictions. For instance, MeSAM \cite{10522788} embeds a novel adapter within the SAM encoder to improve the high-frequency feature for RS optical images. RSAM-Seg \cite{zhang2024rsam} enriches the high-frequency information using residual network connections. SCD-SAM \cite{mei2024scd} introduces a dual-encoder structure to extract semantic change features while incorporating local information. \cite{osco2023segment} combines the specific category tasks with textual prompts to generate general examples. CWSAM \cite{pu2024classwise} employs a class-wise mask decoder to increase low-frequency information in SAR images. Rsprompter \cite{chen2024rsprompter} leverages anchor-based prompt generators and mask decoders to address category-specific inputs, for instance, segmentation. SAGFFNet \cite{chen2024self} uses adaptive modules to reduce the dimensionality of hyperspectral images before inputting. SAMRS \cite{wang2024samrs} introduces a large-scale RS segmentation dataset generated from rotated bounding box prompts. UV-SAM \cite{zhang2024uv} utilizes mixed prompt information from small semantic segmentation models to guide SAM in fine-grained boundary detection.

SAM-related techniques around urban applications in RS can be primarily categorized into three key approaches. \textbf{(1) Sample Annotation.} SAM enables rapid, high-precision semi-automated annotation, significantly reducing the time and effort required for manual labeling. \textbf{(2) Fine-Tuning for Specific Tasks.} MAF-SAM \cite{song2024multispectral} introduces a multi-stage adaptation strategy to fine-tune SAM for land cover classification using multispectral images. SAM-Road \cite{hetang2024segment, zhang2024segment, guo2024few} adapts SAM for road segmentation tasks, while water body extraction primarily focuses on fine-tuning the Mask Decoder \cite{moghimi2024comparative, wang2024monitoring}. \textbf{(3) Incorporating Additional Auxiliary as Prompts.} YOLOSc-SAM \cite{gao4879705yolosc} utilizes anchor boxes generated by YOLO as prompt inputs to guide SAM in segmenting farmland regions. SAM is combined with CLIP \cite{rs16050842} for sea ice extraction to achieve precise segmentation and classification of sea ice regions. SPA \cite{zhao2024spa} proposes leveraging existing object detection bounding boxes as prompts in combination with SAM to create new semantic segmentation datasets. RSPrompter \cite{chen2024rsprompter} utilizes a feature aggregator to locate objects within SAM’s encoder and infer their semantic categories. MW-SAM \cite{zhang2024mw} introduces specific wetland features to enhance prompt inputs. SolarSAM \cite{li2024solarsam} applies text-guided semantic segmentation to accurately segment building rooftops.

The exploration of SAM in RS remains far from exhaustive, as current applications primarily serve as supportive tools or rely on underdeveloped fine-tuning strategies tailored for single-task objectives. For achieving robust segmentation in global urban scenarios, several gaps in the current implementation of SAM can be summarized as follows:

\begin{itemize}
\item \textbf{Gap 1 (\textit{cf.} Scaling Effect):} The RS imaging modality causes the same object to exhibit distinct surface semantic characteristics at varying resolutions. Additionally, different object categories are better represented by their spatial distribution and key features at specific spatial scales. As a result, while SAM may be fine-tuned for particular tasks, it often struggles to adapt to scale variations in other tasks, ultimately compromising the accuracy of feature extraction.
\item \textbf{Gap 2 (\textit{cf.} Manual Prompts):} The segmentation performance of the interactive SAM framework heavily relies on the accuracy of the provided prompts. However, in full-area segmentation tasks, high-quality manual annotations are often limited. The sparse prompts typically available may fail to comprehensively cover the entire sample space, leading to reduced algorithmic stability and inconsistent segmentation results.
\end{itemize}

To effectively address the aforementioned challenges, we aim to design a segmentation model with adaptive prompts capable of handling the diversity and heterogeneity of various urban morphologies. Inspired by MRA theory, we investigate the segmentation performance of urban land cover types across multiple resolutions, yielding key observations as follows (visual evidence is shown in Fig. \ref{fig:motivation}: \textit{(1) Limited point and bounding box prompts constrain receptive fields of SAM, which is particularly problematic for elongated targets such as roads and water bodies, often leading to significant omissions. (2) From the segmentation results at multi-resolution views, consistent with MRA theory, while certain omissions or false alarms may occur within internal structures, inherent invariance across multiple views is observed. Notably, the ratio of the receptive field in the target region relative to the entire image remains stable.} Based on these findings, we propose a flexible adapter designed to learn scale-invariant features across multiple resolutions. Moving away from conventional convolutional flows commonly used in existing methods, we introduce a series of MRA-compliant U-shaped adapters tailored to address challenges arising from varying resolutions, scales, and shapes. Importantly, we incorporate a cross-attention mechanism to align the adapters with the backbone encoder, enabling the transfer of invariant learning while replacing traditional manual interactive prompting. Fig. \ref{fig:motivation} illustrates the design concept of the proposed UrbanSAM, highlighting its significant contributions to addressing scale effects and improving segmentation performance in complex urban environments.
\begin{itemize}
    \item \textbf{Customized SAM for Global Urban Scenes.} To the best of our knowledge, the proposed UrbanSAM is the first foundational model, which is customized for object extraction and segmentation in urban construction. By learning invariant properties across multiple scales, UrbanSAM effectively eliminates the reliance on manual interactive prompts, achieving robust and accurate segmentation of diverse urban features.
    
    \item \textbf{Invariance-Inspired Adapters.}
    Inspired by MSA theory, UrbanSAM employs multiple U-scaling adapters to capture latent commonalities—inherent properties that remain invariant—across diverse urban morphologies at multiple scales. This enables robust and consistent feature representations, effectively addressing global variations in urban environments.
    
    \item \textbf{Injected Domain-Specific Priors.} According to the principle of hierarchical inheritance, UrbanSAM incorporates a cross-attention mechanism to systematically align multiple adapters with the backbone encoder at different stages. This integration equips the encoder with enhanced capabilities for learning invariance, further yielding learnable prompts. This approach not only enhances the robustness of segmentation outcomes but also eliminates the reliance on manual prompts, making the system more efficient and adaptable to diverse urban environments.
 
    \item \textbf{Multi-Scale Accommodators.} UrbanSAM features a regulator that performs sampling operations to accommodate input images of arbitrary dimensions. Additionally, the U-shaped adapters effectively capture comprehensive contextual semantic information at specific resolutions, mitigating accuracy loss caused by scale adjustments and enabling more precise and complete feature extraction across varying spatial scales.
    
    \item \textbf{Superior Performance.} UrbanSAM achieves outstanding performance across diverse urban object extraction tasks, such as building, water body, and road extraction. By consistently outperforming existing state-of-the-art methods, it showcases exceptional effectiveness, robustness, and generalizability in addressing the complexities of urban environments.
\end{itemize}

\section{Invariance Learning Inspired by Multi-Resolution Analysis}

\subsection{Multi-Resolution Analysis Theory}\label{MRA}
From the perspective of computer vision, directly analyzing global image information based solely on individual pixel values presents significant challenges. Instead, greater emphasis is placed on local changes, as these regions typically correspond to key features of the objects of interest. To avoid redundant computations caused by oversaturation, a more comprehensive analysis can only be achieved by determining an optimal scale or resolution. However, the large variation in object sizes within an image complicates the predefinition of a single, universally applicable resolution for analysis. An adaptive multi-scale approach addresses this issue by dynamically adjusting to varying object sizes and resolutions, thereby enabling more accurate analysis.

\begin{figure*}[!t]
      \centering
	   \includegraphics[width=1.0\textwidth]{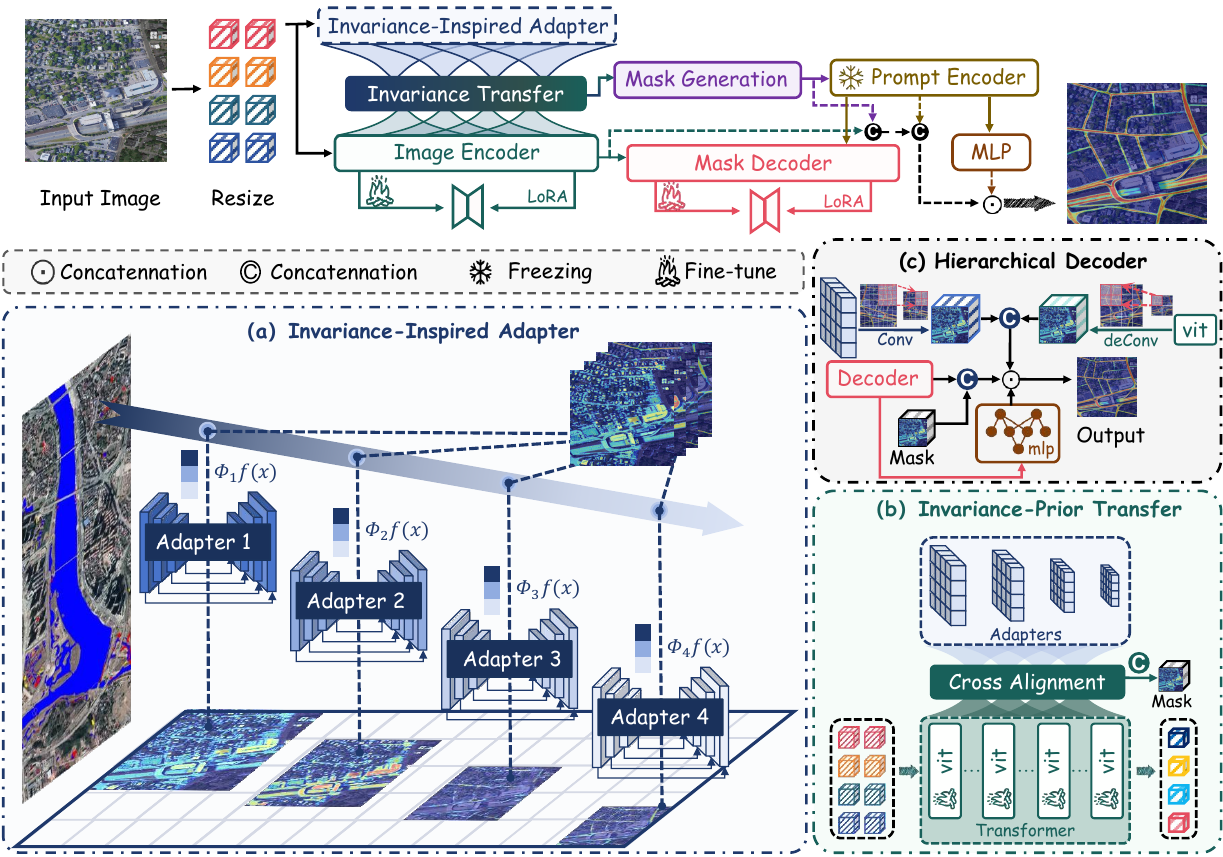}
      \caption{An illustrative workflow of the proposed UrbanSAM, which can segment dominant urban elements (e.g., buildings, water, roads) by leveraging the invariance attribute from MRA theory. (a) Invariance is embedded into the prompter design through cascaded U-scaling adapters (Fig. \ref{fig:adapter}) that capture hidden cues across multiple scales. (b) These learned multiscale features in the prompt stream are transferred to the SAM main body (i.e., transformer blocks), guiding the alignment of subsequent representations and leading to more robust, superior segmentation performance in urban scenes.}
\label{fig:workflow}
\end{figure*} 

The concept of MRA was defined in \cite{mallat1989theory}, which theoretically proves the invariance across different scales. The theory guarantee provides an important basis and source for multi-resolution image representations.

\begin{itemize}
\item $\textbf{Invariance.}$\; Existing an optimal bases $V_{j}$, consisted and translated by the function $V_{0}$.

$\mathit{Theorem\;1}\;\text{(Rieze\;Bases)}$: Let $\mathbf{V}_{j},{j\in \mathbf{Z}}$ be the multi-resolution approximation of $ L^{2}(\mathbb{R})$. The scaling function $\Phi  \left ( x \right ) \in L ^{2} \left ( \mathbb{R} \right )$ satisfies the following condition:
\begin{equation}
    \lim _{j \rightarrow\infty}\left\|f\left ( x \right )-\Phi_{j}f\left ( x \right )\right\|^{2}=0,
\end{equation}
where the function sequence $\left\{\Phi_{j}f\left ( x \right )\right\}_{0}^{j=\infty}$ converges to $f\left ( x \right )$ when $j\to \infty$. 

\item \textbf{Approximation.}\; Assuming that the resolution $2^{j+1}$ contains all the information at the lower resolution $2^{j}$:
\begin{equation}
    V_{j-1} \subset V_{j} \quad \forall j \in \mathbf{Z},
\end{equation}
where $f\left(x\right) \in L^{2}(R)$ is an observational function with $j$ subspace sequence $\left\{V_{j}, j \in \mathbf{Z}\right\}$.

$\mathit{Theorem\;2}$: the approximation operations are similar at all different resolutions, indicating that target sub-space can be deduced or characterized by other scale approximation:
\begin{equation}\label{eq3}
    f(x) \in V_{j-1} \Leftrightarrow f\left(2^{(j-1)}x\right) \in V_{j}.
\end{equation}
\end{itemize}

MRA asserts that any complex image space can be expressed as the product of a set of coefficients $\Phi_{j}$ and an optimal basis space $f\left ( x \right )$, where the basis can be flexibly scaled to accommodate any resolution. This theory implies that for every image, there exists a resolution-invariant basis space that encapsulates the global intrinsic structural properties of the image, remaining unchanged regardless of resolution variations. Consequently, the ability to derive this basis space provides a robust approach to addressing the challenges posed by global image diversity.

\begin{figure*}[!t]
      \centering
	   \includegraphics[width=1.0\textwidth]{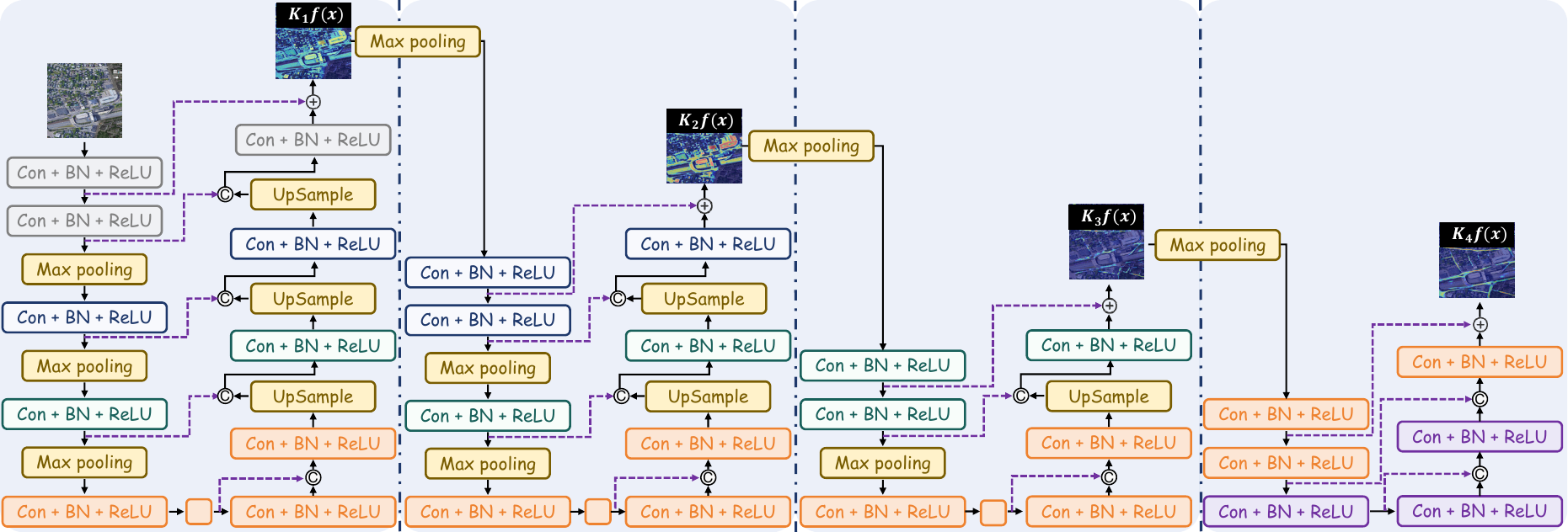}
      \caption{An illustrative workflow of the proposed U-Scaling adapters, which follows MRA theory and aims to provide adaptive and effective prompt guidance for SAM by learning scene intrinsic scale invariance across multiple resolutions, where each adapter is designed to approximate different forms of optimal receptive fields at different resolutions.}
\label{fig:adapter}
\end{figure*} 

\subsection{U-Scaling: Constituting a Refined Adapter for Perceptual Invariance}
The convolution theorem indicates that the convolution operation is equivalent to multiplication in the frequency domain of a signal, a principle that extends to image processing. Consequently, a common approach involves directly or simplistically using a convolutional stream to connect corresponding layers across two paths. However,  Urban RS images are typically characterized by complex semantic scenes, further complicating analysis. Additionally, the substantial scale variation in RS images—from 1000m down to 0.5m ground sampling distances (GSD)—makes single-scale image analysis both challenging and insufficient. This scale diversity often prevents capturing essential details or fully reflecting the semantic context in practical urban scenarios.

The U-shaped encoder is designed as a top-down concatenation of multi-scale features, spanning from shallow layers $\{\Phi_{j}, j=1 \;|\;j=1,2,...,n\}$ to deep layers $\{\Phi_{j}, j=n \;|\;j=1,2,...,n\}$. In contrast, the decoder aggregates multi-scale features from deep layers $\{\Phi_{j}, j=n \;|\;j=1,2,...,n\}$ to shallow layers $\{\Phi_{j}, j=1 \;|\;j=1,2,...,n\}$ in a bottom-up fashion, thereby correcting and enhancing the feature representation capacity. Undoubtedly, this aligns with the invariance learning approximation described in Eq. (\ref{eq3}). As established by the theorem, throughout this scalable process, the expression of the resolution basis space will be involved, allowing the model to learn intrinsic image properties that do not vary with resolution, irrespective of the input image size.

This naturally motivates us to model the scale characteristics of urban RS images by means of MRA. More specifically, we aim to eliminate the dependency on manual prompts by learning the inherent scale invariance of scenes across multiple resolutions, thus enabling adaptive and effective prompt guidance for SAM. We propose an adapter composed of four U-Scaling modules (shown as Fig. \ref{fig:adapter}), where each U-Scaling module is designed to approximate the optimal base space for various forms across different resolutions. This configuration allows for effective segmentation of targets at any resolution and form. By concatenating all adapters and applying weighted adjustments, our method dynamically addresses the challenges posed by the heterogeneity of global urban structures.

Specifically, when the image features $F(x)$ are fed into a U-Scaling module within the adapter, they first undergo two convolutional mappings. Simultaneously, the application of ReLU activation facilitates a nonlinear transformation of the feature space at higher resolutions. Subsequently, a MaxPooling operation is employed to effectively leverage contextual information, achieving a seamless transition across different resolutions within a small stride range. Ultimately, features at the same scale within the U-Scaling module are aggregated via residual connections, thereby enhancing the completeness of feature representation. This expression can be formally represented as:
\begin{equation}
    \begin{aligned}
     \label{eq4}
        & F{(x_{i})} = En_{\Phi_{i,j} f{(x_{i,j})}} + De_{\Phi_{i,j} f{(x_{i,3-j})}} \\
        &~~~~~~~~~~ = \sum_{j=0}^{3} \Phi_{i,j} f{(x_{i,j})} + f{(x_{i,0})},\\
    \end{aligned}
\end{equation}
where $En (\cdot)$ and $De (\cdot)$ represent the feature calculation results of the encoder and decoder, respectively. $\Phi_{i,j}$ represents the mapping coefficient at the $j$-th scale transformation of the $i$-th adapter, with $\Phi$ set to 2 in the experiments. $f{(x_{i,0})}$ denotes the input feature of the $i$-th adapter.

\section{UrbanSAM: Customized SAM for Urban}\label{method}
In this section, we systematically introduce the proposed UrbanSAM, as illustrated in Fig. \ref{fig:workflow}, providing a detailed overview of its design and functionalities. Initially, an adapter composed of multiple U-Scaling modules, inspired by the MRA theory, serves as a critical bridge. A cross-alignment mechanism facilitates the transfer of invariance learning capabilities to the image encoder of UrbanSAM and integrates them into the learning process of the LoRA parameters. Subsequently, invariant intrinsic features across multiple resolutions generate mask prompts, eliminating reliance on manually crafted prompts. Notably, we employ LoRA technology to freeze the core parameters of the transformer blocks within SAM, while only training the additional LoRA parameters. This approach enhances the generalization ability of UrbanSAM and mitigates biases induced by varying data distributions. Further details are provided below.

\subsection{Invariance Transfer}
By employing cross-alignment, the invariance learning capabilities of the adapter are transferred to the image encoder. Concurrently, the multi-scale mask prediction results are utilized to constrain the attention regions of the prompt generator, thereby enhancing the feature representation capacity of the adaptive prompt generator.

\textbf{Cross-alignment mechanism.} \; 
This section primarily utilizes the cross-branch masked attention operator at different scales of U-Scaling output \( F_u \) and ViT through the feature output \( F_v \) after the global attention block and obtained the fusion output \( F \). The specific formula is as follows:
\begin{equation}
\label{eq5}
    \begin{aligned}
    F&(F_v, F_u) = \\
    &M \odot \text{\textit{Softmax}}\left(\frac{(F_v M_q) (F_u M_k)^T)}{\sqrt{d_c}}\right) (F_u M_v) + F_v,
    \end{aligned}
\end{equation}
where \( \text{\textit{Softmax}}(\cdot) \) represents the Softmax function. \( M_q, M_k, M_v \in \mathbb{R}^{d \times d_c} \) correspond to the parameter matrices of Query, Key, Value, mapping the feature dimensions from \( d \) to \( d_c \), where \( d_c \) refer to the dimension of the cross-branch masked attention module, and \( M \) represents the foreground probability of the prediction result obtained through the sigmoid function. After sampling adjustment, the foreground probability is multiplied by the element with the weight matrix obtained through softmax.

\textbf{Low-Rank Adaptation (LoRA).} \;  
In the LoRA-based mapping layer, we denote the weight matrix of the original projection layer as $\mathbf{W}$, which is kept frozen. A low-rank approximation is achieved via a shortcut connection comprising two linear layers. The weight matrices of these layers are represented as \( A \in \mathbb{R}^{r \times C_{\text{in}}} \) and \( B \in \mathbb{R}^{C_{\text{out}} \times r} \), \( r \ll \min(C_{\text{in}}, C_{\text{out}}) \). Matrix \( A \) is initialized with a random Gaussian distribution, while \( B \) is initialized to zero, ensuring that the initial value of \( BA \) is zero. Thus, the modified projection layer is expressed as follows:
\begin{equation}
    \begin{aligned}
        \left\{\begin{aligned} 
        &{Q}_{i}={W}_{q}x + {W}_{Q} {A}_{q}x,\\
        &{K}_{i}={W}_{k}x,\\
        &{V}_{i}={W}_{v}x + {B}_{v} {A}_{v}x,  
        \end{aligned}\right.\\ 
\end{aligned}
\end{equation}
where \(W_q\), \(W_k\), and \(W_v\) are the weight matrices in the frozen original SAM, and \(B_q\), \(A_q\), \(B_v\), and \(A_v\) are the learnable LoRA parameters.

\begin{figure*}[!t]
      \centering
	   \includegraphics[width=1.0\textwidth]{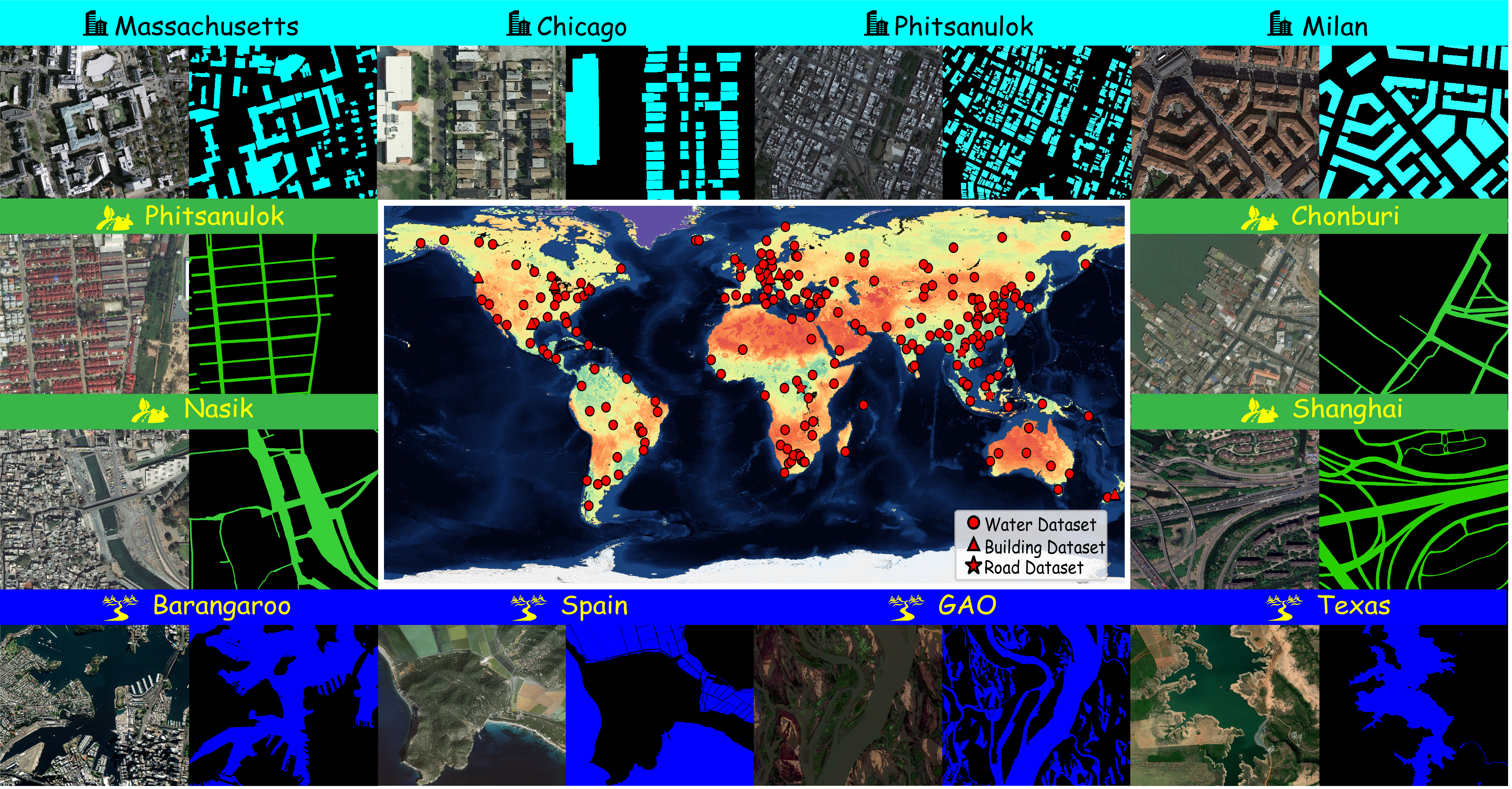}
      \caption{Global distribution of sample datasets for urban construction (building, water, and road), marked by distinct shapes. Paired examples (RS images and corresponding labels) illustrate the content of each dataset.}
\label{fig:datashow}
\end{figure*}

\subsection{Adaptive Mask Generation}
The outputs from multiple U-Scaling modules in the adapter are employed to enhance foreground region predictions, generating accurate prompts. Specifically, multi-scale masks, cross-aligned with the image encoder, are concatenated. A $1 \times 1$ convolutional layer is then applied to predict the target region, with foreground segmentation constrained by binary cross-entropy and Dice loss. The predicted results are subsequently transformed into foreground probabilities using the Sigmoid function, and a learnable parameter, initialized at 0.5, is introduced for filtering. This process can be expressed as follows:
\[
\text{\textit{Mask~prompt}} = 
\begin{cases} 
1, & \text{if} \;\; \sigma(P_{\text{mask}}) \geq \tau \\
0, & \text{otherwise}
\end{cases},
\]
where \( \tau \) is the threshold obtained from network adaptive learning, \( \sigma(\cdot) \) the sigmoid function and \( P_{\text{mask}} \) the prediction result. With this approach, the proposed Learnable Prompt Adapter module can adaptively customize intensive prompts according to different object extraction tasks, eliminating the need for UrbanSAM manual annotation and thus significantly improving UrbanSAM adaptation in complex scenarios.

\subsection{Hierarchical Consistency Decoder}\label{Image Decoder}
The decoder is customized with an additional hierarchical upsampling module, which progressively fuses the intermediate outputs of the image encoder and the adapter through skip connections, serving as both constraints and guidance. This integration effectively incorporates the invariance learning capabilities of the adapter into the decoder. Furthermore, the progressive upsampling strategy, characterized by multiple iterations and small step sizes, better preserves spatial details and class mask information, thereby significantly enhancing the quality of mask generation. Additionally, we introduce a three-layer Multi-Layer Perceptron (MLP) that dynamically generates weights for the original output tokens, which, under deep supervision, include both class and mask information. These tokens are then multiplied spatially with the fused features to produce high-quality mask generation. 

This progressive strategy involves four features $\{H_{i}\;|\;i=1,2,3,4\}$ 1) the first block interacting with the adapter within the ViT backbone encoder ${F_{v}}$; 2) the feature from the U-Scaling module ${F_{u}}$; 3) the mask prediction results ${H}_{1} = {M_{pre}}$; 4) the final output from decoder ${H}_{2} = M_{F_{v}}$. This process can be expressed as:
\begin{equation}\label{eq7}
    \begin{aligned}
        \left\{\begin{aligned} 
        &{H}_{1}= Concat \left (\mathcal{C}^{T}(F_{v}), \; \mathcal{C}(F_{u}) \right),\\
        &{H}_{2}= Concat \left ({H}_{1}, \; \mathcal{C}^{T}(M_{F_{v}}) \right),\\
        &{H}_{3}= Concat \left ({H}_{1}, \; {H}_{2}, \;  {H}_{3}, \; {H}_{4}\right),\\
        &Seg_{Out} = {H}_{3}\odot MLP\big ( M_{F_{v}} \big ),\\
        \end{aligned}\right.\\ 
\end{aligned}
\end{equation}
where $Seg_{Out}$ denotes the final output result, and $\mathcal{C}(\cdot)$ and $\mathcal{C}^{T}(\cdot)$ represent the convolution and transposed convolution operations, respectively. $\odot$ denotes the spatial pointwise multiplication, with dynamic weight generation for tokens achieved through a three-layer MLP operation.

\subsection{Composite Loss Function}
UrbanSAM combines pixel and region-wise loss to jointly optimize segmented predictions and mask prompts. This method enhances the model's grasp of remote sensing structures, improving its real-world performance. The loss function is defined as
\begin{equation}
\label{eq8}
\begin{aligned}
L = \mathit{\lambda}_{bce} {L}_{bce} + \mathit{\lambda}_{dice} {L}_{dice},
\end{aligned}
\end{equation}
where ${L}_{bce}$ and ${L}_{dice}$ represent the binary cross-entropy loss and Dice loss, respectively. $\mathit{\lambda}_{bce}$ and $\mathit{\lambda}_{dice}$ are the weighting parameters used to balance the influence of these two loss components.

Considering the mask supervision in mask cross-attention and mask prompt learning, as well as the deep supervision strategy applied in the hierarchical consistency mask decoder structure, the final loss function is composed of three components, as specified by the following formula
\begin{equation}
\label{eq9}
\begin{aligned}
L_{all} = L\left (\hat{y}, y \right) + L\left( \widehat{y_{\frac{1}{4}}}, D\left( y \right)\right) + \frac{1}{n}\sum_{i=1}^{n}L\left( \widehat{y^{n}_{mask}}, D\left( y \right)\right).
\end{aligned}
\end{equation}
In this context, $y$ denotes the ground truth, $\hat{y}$ represents UrbanSAM's final prediction, and $\widehat{{y}_{1/4}}$ refers to the deep supervision prediction at 1/4 resolution. $\widehat{{y}^{n}_{mask}}$ represents the output of the $n$-th mask, where $n$ is set to 5 considering the mask prompt.

\section{Experiments}  \label{experiments}
\subsection{Experimental Setting}
Five commonly used evaluation metrics, Overall Accuracy (OA), Precision, Recall, F1, and Intersection over Union (IoU), are employed to assess the segmentation performance of UrbanSAM and the comparison methods. Specifically, OA represents the percentage of correctly classified pixels relative to the total number of pixels. Precision quantifies the proportion of correctly predicted positive samples among all instances predicted as positive. At the same time, Recall measures the ratio of correctly predicted positive samples to the total number of positive samples in the ground truth data. The F1 is the harmonic mean of Precision and Recall, providing a balanced measure of both metrics. Finally, IoU is defined as the ratio of the intersection to the union of the predicted and ground truth (GT) regions. 

Fig. \ref{fig:datashow} further demonstrates the geographic coverage of the dataset and details its wide distribution worldwide (including Asia, Oceania, North America, Europe, and Africa). This highlights differences in the distribution of data across regions, further demonstrating the broad applicability of the model to adapt to different geographical settings.

The experimentation uses the PyTorch framework, with training performed on four NVIDIA L40 GPUs, each equipped with 48GB of memory. The training loss function is defined as a weighted combination of cross-entropy loss and Dice loss, with the weight for cross-entropy set to 0.2 and for Dice loss set to 0.8. The model is optimized using the Stochastic Gradient Descent (SGD) optimizer, with momentum and weight decay values of 0.9 and 0.0001, respectively. The learning rate (lr) is adjusted by combining a warmup phase and an exponential decay strategy. The LoRA method is employed for fine-tuning only the frozen \textit{q} and value \textit{v} projection layers within the transformer blocks. The LoRA rank is set to 4 to optimize efficiency and performance, with the hyperparameter alpha configured as twice the rank.

\begin{figure*}[!t]
      \centering
	   \includegraphics[width=1.0\textwidth]{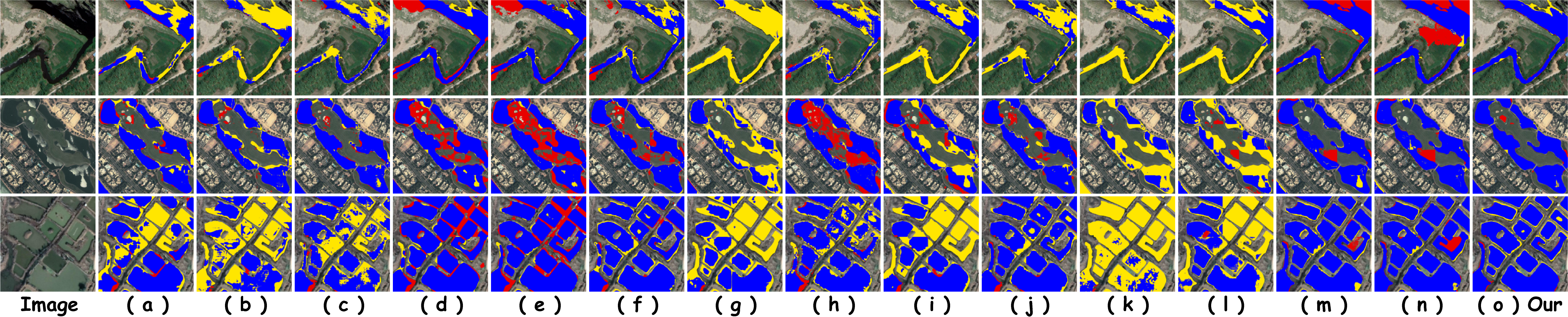}
      \caption{Visualization of water body extraction results using various methods compared to our UrbanSAM, where red represents false positives, and yellow indicates false negatives. (A)-(C) present large-scale qualitative outcomes from UrbanSAM across three regions. (a)–(o) illustrate water body extraction in selected ROIs using FCN8s, UNet, LinkNet50, PSPNet, DeepLabv3+, HRNetv2, MECNet, Segformer, Uformer, MFSegformer, SAM, SAMDB, HQSAMDB, and UrbanSAM, respectively.}
\label{fig:water}
\end{figure*}
\begin{table}[!t]
\centering
\caption{Quantitative comparison between advanced methods and our UrbanSAM for water body extraction in urban construction. Evaluation metrics include OA, Precision, Recall, F$_{1}$ score, and IoU. The highest value in each metric is shown in bold.}
\resizebox{0.5\textwidth}{!}{
\begin{tabular}{cccccc}
\toprule[1.0pt]
\multirow{2}{*}{{Method}} & \multicolumn{5}{c}{{Water}} \\  
\cmidrule(r){2-6}  
 & {OA} & {Precision} & {Recall} & F$_1$ & {IoU} \\
\midrule
\midrule
FCN8s & 92.80 & 93.19 & 80.65 & 86.46 & 76.16\\
        UNet & 89.18 & 87.60 & 72.31 & 79.23 & 65.60\\
        LinkNet50 & 93.01 & 93.14 & 81.50 & 86.93 & 76.88\\
        PSPNet & 93.09 & 90.17 & 85.05 & 87.53 & 77.83\\
        DeepLabv3+ & 93.47 & 89.94 & 86.82 & 88.35 & 79.14\\
        HRNetv2 & 94.02 & 92.29 & 86.26 & 89.17 & 80.46\\
        MECNet & 88.41 & 93.94 & 63.46 & 75.75 & 60.97\\
        MSResNet & 92.81 & 89.83 & 84.36 & 87.00 & 77.00\\
        \midrule
        Segformer & 90.06 & 88.18 & 75.24 & 81.20 & 68.34\\
        Uformer & 91.90 & 90.44 & 80.06 & 84.93 & 73.81\\
        MFSegformer & 84.79 & 81.84 & 59.98 & 69.23 & 52.94\\
        SAM (B) & 89.41 & 89.56 & 71.19 & 79.32 & 65.73\\
        SAM (L) & 87.73 & 87.04 & 66.95 & 75.68 & 60.88\\
        SAM (H) & 88.20 & 90.68 & 65.35 & 75.96 & 61.23\\
        SAMDB (H) & 94.82 & 92.08 &  \textbf{89.54} & \underline{90.79} & 83.13\\
        HQSAMDB (H) & 93.97 & 90.04 & \underline{88.68} & 89.35 & 80.76\\
        \midrule\midrule
        \cellcolor{gray!15}UrbanSAM (B) & \cellcolor{gray!15}94.31 & \cellcolor{gray!15}94.76 & \cellcolor{gray!15}84.75 & \cellcolor{gray!15}89.48 & \cellcolor{gray!15}80.96\\
       \cellcolor{gray!35} UrbanSAM (L) & \cellcolor{gray!35}\underline{94.96} & \cellcolor{gray!35}\underline{95.26} & \cellcolor{gray!35}86.65 & \cellcolor{gray!35}90.75 & \cellcolor{gray!35}\underline{83.07}\\
        \cellcolor{gray!60}UrbanSAM (H) &  \cellcolor{gray!60}\textbf{95.20} &  \cellcolor{gray!60}\textbf{95.75} & \cellcolor{gray!60}87.05 &  \cellcolor{gray!60}\textbf{91.19} & \cellcolor{gray!60} \textbf{83.81}\\
\bottomrule[1.0pt]
\end{tabular}
}
\label{tab:water}
\end{table}

To ensure a fair comparison, all methods are evaluated under identical experimental conditions, utilizing official code implementations or pre-trained weights. We strictly adhere to the original hyperparameter configurations for domain-specific methods as specified in the respective publications.

\subsection{Urban Water}
\subsubsection{Dataset}
Regarding the Urban Water task, we used the GLH-Water dataset \cite{li2024glh}. This dataset consists of 250 satellite images from Google Earth, encompassing every continent except Antarctica. Each image has a high resolution of roughly 0.3 meters, spans an area of 3,686 km$^2$, and measures 12,800 by 12,800 pixels. In line with the dataset’s original partitioning scheme, we divided the images into non-overlapping patches of 512$\times$512 pixels, which yielded 125,000 training samples and 7,319 testing samples.

\subsubsection{Implementation details}
In the experiments, we set the input image size to 512$\times$512 to retain as much of the integrity and continuity of water bodies in the remote sensing imagery as possible. The initial LR for UrbanSAM is set to 0.001 without LR scheduling strategies. Following the original configurations outlined in the dataset paper, the proposed model and the comparison methods are trained for 15 epochs on each dataset without data augmentation. 

\begin{figure*}[!t]
      \centering
	   \includegraphics[width=1.0\textwidth]{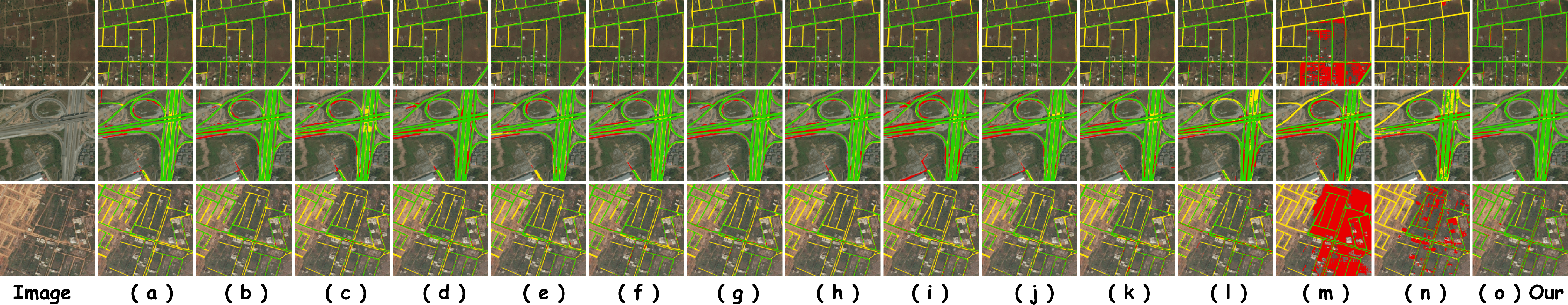}
      \caption{Visualization of road extraction results using various methods compared to our UrbanSAM, where red represents false positives and yellow indicates false negatives. (A) presents a large-scale qualitative map in the USA generated by UrbanSAM. (a)–(o) show road extraction in selected ROIs using FCN8s, UNet, LinkNet50, PSPNet, DeepLabv3+, HRNetv2, MECNet, Segformer, Uformer, MFSegformer, SAM, SAMDB, HQSAMDB, and UrbanSAM, respectively.}
\label{fig:road}
\end{figure*}
\begin{table}[!t]
\centering
\caption{Quantitative comparison between advanced methods and our UrbanSAM for road extraction in urban construction. Evaluation metrics include OA, Precision, Recall, F$_{1}$ score, and IoU. The highest value in each metric is shown in bold.}
\resizebox{0.5\textwidth}{!}{
\begin{tabular}{cccccc}
\toprule[1.0pt]
\multirow{2}{*}{{Method}} & \multicolumn{5}{c}{{Road}} \\  
\cmidrule(r){2-6}  
 & {OA} & {Precision} & {Recall} & F$_1$ & {IoU} \\
\midrule
\midrule
{U-Net}  & 97.38 & 84.48 & 66.25 & 74.26 & 59.06  \\
LinkNet50 & 97.64 & 83.16 & 73.50 & 78.03 & 63.98  \\
{DlinkNet50} & 97.37 & \textbf{84.88} & 65.56 & 73.98 & 58.71 \\
{Deeplabv3+} & 97.63 & 82.00 & 74.84 & 78.26 & 64.28 \\
{RoadCNN} & 97.67 & 83.95 & 73.25 & 78.24 & 64.25 \\
{GAMSNet} & 97.58 & 81.59 & 74.41 & 77.83 & 63.71 \\
{CoANet}  & 97.55 & 83.34 & 71.25 & 76.82 & 62.37 \\
{MSMDFF} & 97.56 & \underline{84.82} & 69.73 & 76.54 & 61.99 \\ 
\midrule
{Segformer} & 97.19 & 77.43 & 71.74 & 74.48 & 59.33 \\
{Uformer} & 97.28 & 78.73 & 71.64 & 75.02 & 60.03 \\
{PVT-UNet} & 96.87 & 77.82 & 63.08 & 69.68 & 53.47 \\
\midrule
{SAM (B)} & 96.24 & 73.25 & 53.69 & 62.13 & 44.89 \\
SAM (L) & 96.16 & 72.75 & 52.20 & 60.78 & 43.66 \\
SAM (H) & 95.91 & 70.08 & 49.51 & 58.03 & 40.87 \\
SAMDB (H) & 79.06 & 16.34 & 64.76 & 26.10 & 15.01  \\
HQSAMDB (H)  & 91.18 & 30.73 & 43.43 & 35.99 & 21.94 \\
\midrule\midrule
\cellcolor{gray!15}{UrbanSAM (B)} & \cellcolor{gray!15}97.77 & \cellcolor{gray!15}80.36 & \cellcolor{gray!15}80.55 & \cellcolor{gray!15}80.46 & \cellcolor{gray!15}67.30 \\
\cellcolor{gray!35}UrbanSAM (L) & \cellcolor{gray!35}\underline{97.93} & \cellcolor{gray!35}\underline{82.11} & \cellcolor{gray!35}\underline{81.56} & \cellcolor{gray!35}\underline{81.83} & \cellcolor{gray!35}\underline{69.25} \\
\cellcolor{gray!60}UrbanSAM (H) & \cellcolor{gray!60}\textbf{97.95} & \cellcolor{gray!60}\textbf{82.18} & \cellcolor{gray!60}\textbf{81.90} & \cellcolor{gray!60}\textbf{82.04} & \cellcolor{gray!60}\textbf{69.55} \\
\bottomrule[1.0pt]
\end{tabular}
}
\label{tab:road}
\end{table}

\subsubsection{Quantitative and Qualitative Comparisons}
Tab. \ref{tab:water} presents the quantitative evaluation results of UrbanSAM in the water body extraction task, demonstrating its superior performance. Its ViT-H achieved an IoU of 83.81$\%$ without human prompts, significantly outperforming the best baseline method by 3.35$\%$. This improvement is attributed to the multiscale invariant features provided by the UrbanSAM prompting strategy, which enables the model to move beyond sole reliance on the color differences between water bodies and adjacent surfaces, instead integrating shape information. This fusion enhances the model's generalization ability for large-scale water body extraction in scenarios where color and shape features exhibit instability. 

Fig. \ref{fig:water} (A) illustrates the remarkable generalization capability of UrbanSAM in large-scale water body extraction, accurately capturing water bodies of various shapes across diverse geographic landscapes and complex land cover conditions. Compared with other SOTA methods, the first and second rows demonstrate adaptability to complex scenarios, effectively avoiding interference from similar land cover, thereby significantly reducing omissions and false detections. The third row highlights the ability to finely segment water bodies at different scales. Furthermore, UrbanSAM excels in capturing the fine details of water body edges, indicating its great potential for detailed urban water mapping on a global scale.

\subsection{Urban Road}
\subsubsection{Dataset}
For the Urban Road task, we conducted experiments using the DeepGlobe \cite{demir2018deepglobe} and LSRV \cite{lu2021gamsnet} datasets. The DeepGlobe dataset is sourced from the DigitalGlobe + Vivid Images collection, featuring areas across Thailand, Indonesia, and India, totaling 2,220 km$^2$. It comprises 6,226 annotated images, each measuring 1024$\times$1024 pixels with a spatial resolution of 0.5 m, covering three RGB bands. For our experiments, we allocated 4,980 images for training and 1,246 for testing. The LSRV dataset includes accurately labeled satellite imagery from Google Earth, covering Boston (USA), Birmingham (UK), and Shanghai (China), with spatial resolutions ranging from 0.36m to 0.51m. The images were processed into non-overlapping patches of 1024$\times$1024 pixels, resulting in 1,041 training and 261 testing samples.

\begin{figure*}[!t]
      \centering
	   \includegraphics[width=1.0\textwidth]{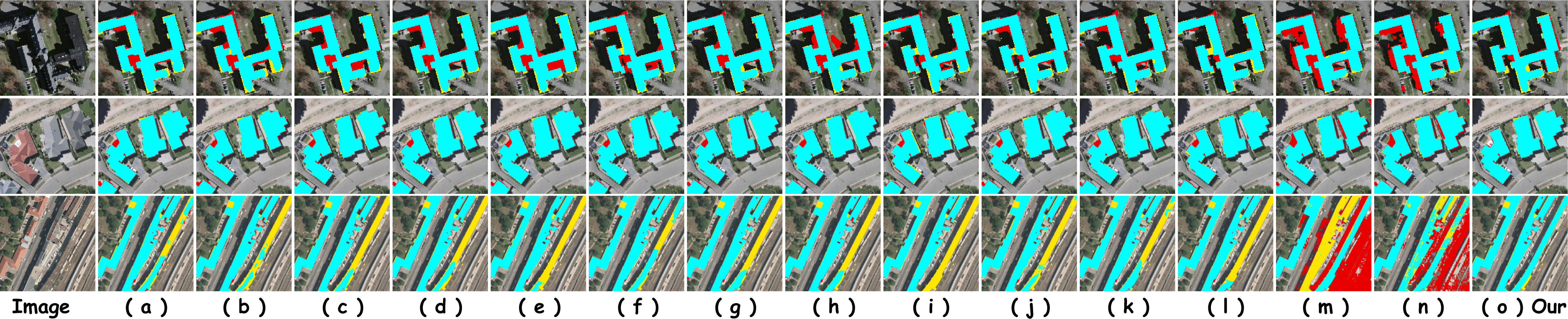}
      \caption{Visualization of building extraction results using various methods compared to our UrbanSAM, where red represents false positives and yellow indicates false negatives. (A) shows a large-scale qualitative map in New Zealand generated by UrbanSAM. (a)-(o) illustrate building extraction in selected ROIs using FCN8s, UNet, LinkNet50, PSPNet, DeepLabv3+, HRNetv2, MECNet, Segformer, Uformer, MFSegformer, SAM, SAMDB, HQSAMDB, and UrbanSAM, respectively.}
\label{fig:build}
\end{figure*}
\begin{table}[!t]
\centering
\caption{Quantitative comparison between advanced methods and our UrbanSAM for building extraction in urban construction. Evaluation metrics include OA, Precision, Recall, F$_{1}$ score, and IoU. The highest value in each metric is shown in bold.}
\resizebox{0.5\textwidth}{!}{
\begin{tabular}{cccccc}
\toprule[1.0pt]
\multirow{2}{*}{{Method}} & \multicolumn{5}{c}{{Building}} \\  
\cmidrule(r){2-6}  
 & {OA} & {Precision} & {Recall} & F$_1$ & {IoU} \\
\midrule
\midrule
UNet & 97.45 & 90.45 & 89.50 & 89.97 & 81.77  \\
DeepLabv3+ & 97.31 & 90.16 & 88.62 & 89.38 & 80.81 \\
HRNetv2 & 97.43 & 90.62 & 89.10 & 89.85 & 81.57\\
U2Net & 97.69 & 92.18 & 89.56 & 90.85 & 83.23 \\
MAPNet & 97.41 & 90.35 & 89.31 & 89.83 & 81.53\\
CBRNet & 97.79 & 92.01 & 90.61 & 91.30 & 84.00 \\
CGSANet & 97.76 & 90.59 & 92.07 & 91.33 & 84.04  \\
HDNet & 97.91 & \underline{92.33} & 91.24 & 91.79 & 84.82\\
\midrule
Segformer & 97.43 & 90.63 & 89.10 & 89.86 & 81.58 \\
Uformer & 97.18 & 89.39 & 88.39 & 88.89 & 80.00 \\
BuildFormer & 97.82 & 92.20 & 90.62 & 91.40 & 84.17 \\
Rsbuilding (B) & 97.84 & 92.32 & 90.65 & 91.48 & 84.29 \\
Rsbuilding (L) & 97.90 & \textbf{92.34} & 91.12 & 91.73 & 84.72\\
SAMDB (H) & 96.92 & 87.81 & 88.10 & 87.96 & 78.51 \\
HQSAMDB (H) & 97.02 & 88.22 & 88.54 & 88.38 & 79.18\\
\midrule \midrule
\cellcolor{gray!15}UrbanSAM (B)& \cellcolor{gray!15}97.85 & \cellcolor{gray!15}91.32 & \cellcolor{gray!15}91.88 & \cellcolor{gray!15}91.60 & \cellcolor{gray!15}84.50  \\
\cellcolor{gray!35}UrbanSAM (L) & \cellcolor{gray!35}\underline{98.00} & \cellcolor{gray!35}92.16 & \cellcolor{gray!35}\underline{92.18} & \cellcolor{gray!35}\underline{92.17} & \cellcolor{gray!35}\underline{85.48}\\
\cellcolor{gray!60}UrbanSAM (H) & \cellcolor{gray!60}\textbf{98.07} & \cellcolor{gray!60}92.15 & \cellcolor{gray!60}\textbf{92.81} & \cellcolor{gray!60}\textbf{92.48} & \cellcolor{gray!60}\textbf{86.01}\\
\bottomrule[1.0pt]
\end{tabular}
}
\label{tab:building}
\end{table}

\subsubsection{Implementation details}
The input for the road extraction task consists of 1024$\times$1024 remote sensing images. Data augmentation techniques, including random rotation and flipping, were applied. UrbanSAM and the comparison methods were trained for 200 epochs, with a warmup strategy employed during the first five epochs. The initial lr was set to 0.005.

\begin{table*}[!t]
\caption{Learning performance of the prompt generator (mask/label, point, box, and our learned prompter). The accuracy of the learned prompt information in UrbanSAM is evaluated by simulating various overlap ratios between the prompt labels and GT. The correctness of mask and box prompts is quantitatively assessed using IoU, while points are evaluated by calculating the matching ratio with GT points. The entries highlighted in bold indicate the best performance across the evaluated metrics, and the unavailable values are denoted by `` - ''.}
    \centering
    \small
    \setlength\tabcolsep{2pt}
    \begin{tabular}{cc}
    {
    \begin{tabular}{cccccc}
    \toprule[1.0pt]
    \multirow{3}{*}{Overlap Ablation}& \multicolumn{5}{c}{{{Mask (Label)}}}  \\     
    \cmidrule(r){2-6}
    & {OA} & {Precision} & {Recall} & F$_1$ & {IoU} \\
    \midrule \midrule 
    100 $\%$ & \textcolor{cyan!100}{\textbf{86.14}} & \textcolor{cyan!100}{\textbf{71.12}} & \textcolor{cyan!100}{\textbf{43.98}} & \textcolor{cyan!100}{\textbf{54.35}} & \textcolor{cyan!100}{\textbf{37.31}} \\ 
    90 $\%$ & \textcolor{cyan!60}{\textbf{85.52}} & \textcolor{cyan!60}{\textbf{69.19}} & \textcolor{cyan!60}{\textbf{41.13}} & \textcolor{cyan!60}{\textbf{51.59}} & \textcolor{cyan!60}{\textbf{34.76}} \\
    70 $\%$ & 83.49 & 60.07 & \textcolor{cyan!30}{\textbf{35.64}} & 44.74 & 28.82 \\
    50 $\%$ & 82.09 & 53.72 & 32.62 & 40.59 & 25.46 \\
    \midrule   \midrule 
    \textbf{Our} & \textcolor{blue}{\textbf{84.40}} & \textcolor{blue}{\textbf{66.17}} & \textcolor{blue}{\textbf{34.40}} & \textcolor{blue}{\textbf{45.26}} & \textcolor{blue}{\textbf{29.25}} \\  
    \bottomrule[1.0pt]
\end{tabular} 
    } &
    {      
    \begin{tabular}{ccccc}
    \toprule[1.0pt]
    \multicolumn{5}{c}{{{Point}}}  \\
    \cmidrule(r){1-5}
    {OA} & {Precision} & {Recall} & F$_1$ & {IoU} \\
    \midrule \midrule 
    {71.61} & {37.93} & \textcolor{cyan!100}{\textbf{80.66}} & \textcolor{cyan!100}{\textbf{51.59}} & \textcolor{cyan!100}{\textbf{34.77}} \\
    63.36 & 31.56 & \textcolor{cyan!60}{\textbf{81.62}} & \textcolor{cyan!60}{\textbf{45.52}} & \textcolor{cyan!60}{\textbf{29.47}} \\
    54.51 & 26.44 & \textcolor{cyan!30}{\textbf{80.01}} & 39.75 & 24.80 \\
    48.40 & 23.72 & 79.01 & {36.48} & 22.31 \\
    \midrule   \midrule 
    - & - & - & - & - \\
    \bottomrule[1.0pt]
\end{tabular}   
    } 
    {      
    \begin{tabular}{ccccc}
\toprule[1.0pt]
    \multicolumn{5}{c}{{{Box}}}  \\
    \cmidrule(r){1-5}
    {OA} &  {P} & {R} & {F$_1$} & {IoU}  \\
    \midrule \midrule 
    36.77 & 21.72 & \textcolor{cyan!100}{\textbf{91.03}} & 35.07 & 21.26 \\
    36.45 & 21.51 & \textcolor{cyan!60}{\textbf{90.18}} & 34.74 & 21.02 \\
    48.61 & 22.36 & \textcolor{cyan!30}{\textbf{70.41}} & 33.95 & 20.44 \\
    71.15 & 22.67 & 22.32 & 22.49 & 12.67 \\
    \midrule\midrule
    - & - & - & - & - \\
\bottomrule[1.0pt]
\end{tabular}
}
\end{tabular}\label{tab:prompter}
\end{table*}

\subsubsection{Quantitative and Qualitative Comparisons}
Tab. \ref{tab:road} shows that all three versions of UrbanSAM achieve SOTA performance in the road extraction task, with IoU improvements of 3.02$\%$, 4.97$\%$, and 5.27$\%$ over the best-performing Deeplabv3+. Furthermore, comparisons with SAM and HQ-SAM highlight that UrbanSAM’s learnable prompting mechanism overcomes the limitations of the traditional box and point-based prompts. Even when handling intricate and elongated road structures in remote sensing images, UrbanSAM effectively reduces false detections and false positives, demonstrating superior adaptability in complex feature extraction tasks.

\begin{figure}[!t]
      \centering
	   \includegraphics[width=0.45\textwidth]{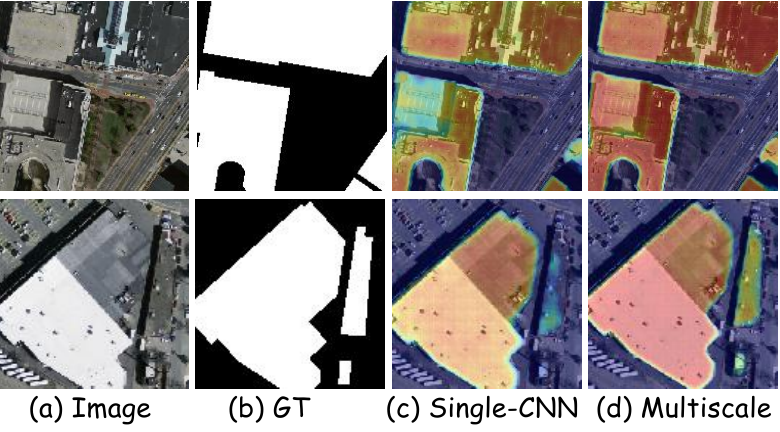}
      \caption{Ablation study on the effectiveness of single-CNN and our multiscale learnable adapter.}
\label{fig:abl_scale}
\end{figure}

Fig. \ref{fig:road} visualizes the precise extraction of continuous and non-fragmented road networks over a large-scale Boston map. The localized scenes Fig. \ref{fig:road} (a)-(o) further emphasize its excellence in capturing structural details. Scene 1 demonstrates its fine extraction capability in cases where roads and backgrounds exhibit similar colors and blurred textures. Scene 2 illustrates its ability to maintain road connectivity in topologically complex looped areas. Scene 3 highlights its superior performance in complex scenes with high spatial heterogeneity.

\subsection{Urban Building}
\subsubsection{Dataset}
Regarding the Urban Building task, we selected three representative datasets: Massachusetts \cite{mnih2013machine}, Inria\cite{maggiori2017can}, and WHU \cite{ji2018fully}. The Boston dataset includes aerial images from the Boston metropolitan area, featuring a spatial resolution of 1m, covering approximately 340 km$^2$. Each 1500$\times$1500 pixel image represents 2.25 km$^2$, showcasing buildings of various scales and architectures. To ensure reproducibility and fairness, the dataset was partitioned as per its original configuration, cropping images and labels into 512$\times$512 patches with 50\% overlap, resulting in 2800 training patches and 250 testing patches. The Inria dataset comprises aerial RGB images from 10 cities in the USA and Austria, spanning 810 km$^2$. According to official guidelines, 5000$\times$5000-pixel images were cropped into 512$\times$512 patches with 1\% overlap, excluding significantly imbalanced samples, yielding 12,315 patches for training and 2500 for validation. Lastly, the WHU dataset, sourced from Christchurch, New Zealand, features a high resolution of 0.3m and an area of over 450 km$^2$. Following the original study’s preprocessing method, non-overlapping 512$\times$512 image patches were created, resulting in a dataset of 8188 patches, with 4736 for training and 2416 for testing.
\begin{figure}[!t]
      \centering
	   \includegraphics[width=0.45\textwidth]{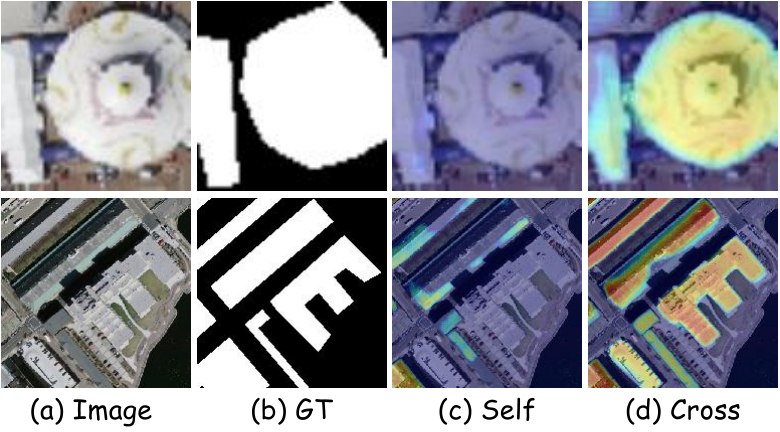}
      \caption{Visualization of class activation maps for self-attention and cross-attention (invariance injection).}
\label{fig:abl_cross}
\end{figure}

\subsubsection{Implementation details}
The input images were uniformly resampled for the building extraction task from 512$\times$512 to 1024$\times$1024. UrbanSAM and the compared methods were trained for 200 epochs, with data augmentation consisting solely of random rotation and flipping. Following the training protocol of the previous SAM model (SAMed), we applied a warmup strategy during the first five epochs, with an initial lr set to 0.005.

\subsubsection{Quantitative and Qualitative Comparisons}
Tab. \ref{tab:building}  presents a quantitative evaluation of UrbanSAM for building extraction, comparing it with SOTA methods. The results demonstrate that UrbanSAM achieves optimal performance in extraction accuracy. Specifically, UrbanSAM outperforms SAM and HQ-SAM, with IoU improvements of 7.5$\%$ and 6.83$\%$, respectively. Unlike RSBuilding, which utilizes learnable query embeddings, UrbanSAM surpasses its performance by 0.21$\%$ and 0.76$\%$ in the ViT-B and ViT-L versions, respectively. This improvement is attributed to UrbanSAM's scale-invariant prompting strategy, which enables the model to more effectively capture the consistent geometric properties of buildings, thereby preserving fine internal details and contours.

\begin{figure*}[!t]
      \centering	   
      \includegraphics[width=0.95\textwidth]{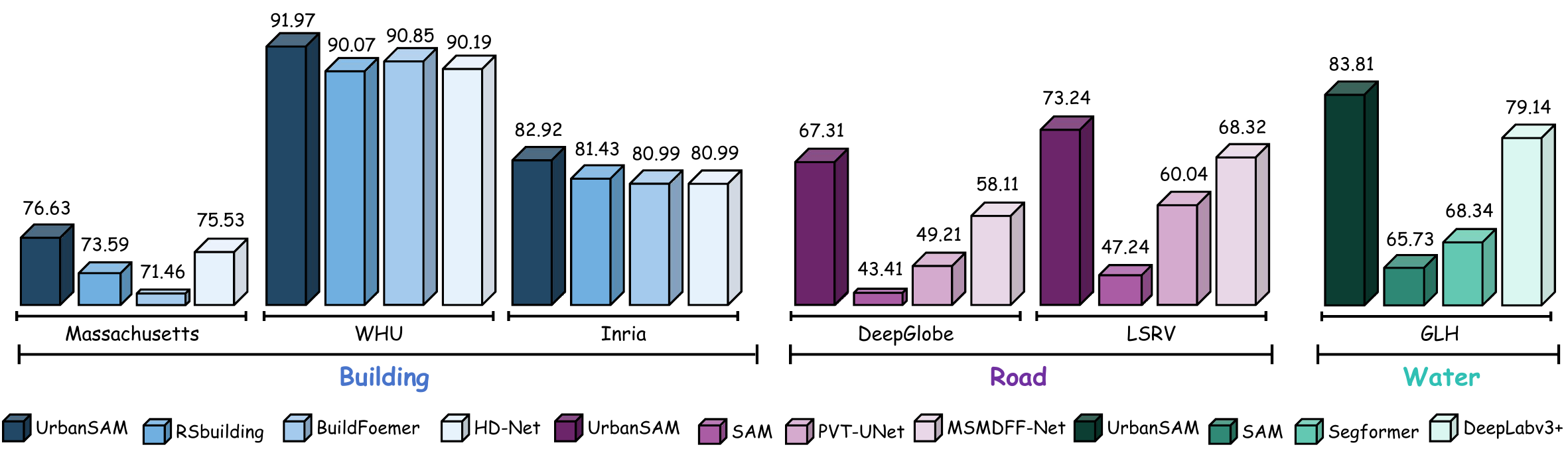}
\caption{Performance comparison of Urban-SAM with other SOTA methods on a global urban segmentation dataset encompassing three buildings, two roads, and one water dataset.}
\label{fig:show_tab}
\end{figure*}
\begin{table*}[!t]
    \centering
   \caption{Ablation Analysis of the proposed UrbanSAM in applying LoRA fine-tuning on different locations and the component combinations on the Massachusetts dataset. The best and second-best results are in blue font and underlined at the bottom. \includegraphics[scale=0.02]{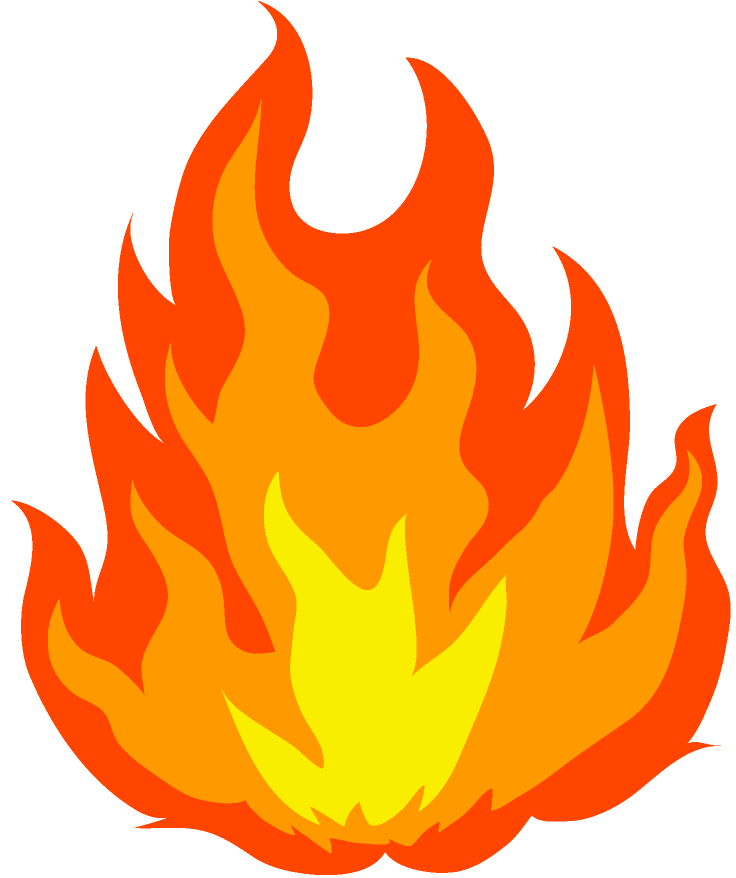} and \includegraphics[scale=0.12]{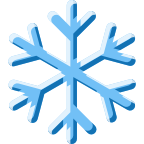} represent freezing and fine-tuning operations.}
    \begin{tabular}{cc}
    \subtable[LoRA Strategy]
    {\resizebox{0.47\textwidth}{!}{
    \begin{tabular}{cc|ccccc}
	\toprule
        {Image} & {Mask} & \multirow{2}{*}{OA} & \multirow{2}{*}{Precision} & \multirow{2}{*}{Recall} & \multirow{2}{*}{F1} & \multirow{2}{*}{IoU} \\
        {Encoder} & {Decoder} & & & & & \\
        \midrule \midrule 
        {\includegraphics[scale=0.12]{figures/FIG/snow.png}} & {\includegraphics[scale=0.02]{figures/FIG/fire.png}} & 93.88 & 82.94 & 85.04 & 83.98 & 72.38 \\
        {\includegraphics[scale=0.12]{figures/FIG/snow.png}} & LoRA &\textbf{\underline{94.09}} & \textbf{\underline{83.35}} & \textbf{\underline{85.83}} & \textbf{\underline{84.57}} & \textbf{\underline{73.26}} \\
        LoRA & {\includegraphics[scale=0.02]{figures/FIG/fire.png}} & 94.04 & 83.21 & 85.67 & 84.43 & 73.05 \\
        LoRA & LoRA & \textcolor{cyan}{\textbf{94.83}} & \textcolor{cyan}{\textbf{86.20}} & \textcolor{cyan}{\textbf{86.27}} & \textcolor{cyan}{\textbf{86.23}} & \textcolor{cyan}{\textbf{75.80}} \\
	\bottomrule 
\end{tabular}   
}
\label{tab:lora}
}
\hfill
\subtable[Different Component Combinations]
   {
   \resizebox{0.5\textwidth}{!}{
   \begin{tabular}{cccccc}
        \toprule 
        {Method} & {OA} & {Precision} & {Recall} & {F1} & {IoU} \\
        \midrule\midrule
        w/o LoRA & 94.48 & 85.04 & 85.82 & 85.43 & 74.56 \textcolor{blue}{(1.24\%↓)} \\
        w/o MultiScale & 94.62 & 85.55 & 85.85 & 85.70 & 74.97 \textcolor{blue}{(0.83\%↓)} \\
        w/o Interaction & 94.75 & 86.26 & 85.65 & 85.96 & 75.37 \textcolor{blue}{(0.43\%↓)} \\
        w/o Decoder & 94.20 & 83.81 & 85.83 & 84.81 & 73.62 \textcolor{blue}{(1.99\%↓)} \\
        \midrule 
        \textbf{UrbanSAM} & \textcolor{cyan}{\textbf{94.83}} & \textcolor{cyan}{\textbf{86.20}} & \textcolor{cyan}{\textbf{86.27}} & \textcolor{cyan}{\textbf{86.23}} & \textbf{\textcolor{cyan}{75.80} }\\
	\bottomrule
\end{tabular}}
\label{tab:component}
    } 
\end{tabular}
\label{ablation}
\end{table*}

Fig. \ref{fig:build} (A) visually demonstrates the ability of our method to accurately capture diverse building structures in complex, large-scale environments, ranging from dense urban areas to scattered suburban buildings. The first scene emphasizes the advantage of understanding regular boundaries, effectively mitigating shadow interference. The second scene highlights its precision in semantic representation, distinguishing interference from similar objects. The third scene illustrates the robust discriminative ability to discern complex, cluttered backgrounds, facilitating the identification of large structures across the entire image. Our method excels in capturing various building forms and demonstrates strong adaptability.

\begin{table*}[!t]
    \centering
   \caption{Ablation study of the LoRA fine-tuning strategy, including the rank size and the impact of applying LoRA at different projection layers on UrbanSAM performance. The best and second-best results are highlighted in bold red and underlined blue, respectively.}
    \begin{tabular}{cc}
    \subtable[The rank size on the LoRA layer.]
    {\resizebox{0.45\textwidth}{!}{
    \begin{tabular}{ccccccc}
	\toprule
        {Rank Size} & {OA} & {Precision} & {Recall} & {F1} & {IoU} \\
        \midrule\midrule
        1 & 94.34 & 85.05 & 84.72 & 84.89 & 73.74 \\
        4 & \textcolor{red}{\textbf{94.83}} & \textcolor{red}{\textbf{86.20}} & \textcolor{blue}{\underline{86.27}} & \textcolor{red}{\textbf{86.23}} & \textcolor{red}{\textbf{75.80}}  \\
        8 & \textcolor{blue}{\underline{94.71}} & 85.18 & \textcolor{red}{\textbf{86.90}} & \textcolor{blue}{\underline{86.04}} & \textcolor{blue}{\underline{75.49}} \\
        16 & 94.58 & \textcolor{blue}{\underline{85.57}} & 85.55 & 85.58 & 74.76 \\
	\bottomrule
\end{tabular}   
}
\label{tab:rank_size}
}
\hfill
\subtable[Different projection layers of LoRA application.]
   {
   \resizebox{0.5\textwidth}{!}{
   \begin{tabular}{cccccc}
        \toprule 
        {Embedding Layer} & {OA} & {Precision} & {Recall} & {F1} & {IoU} \\
        \midrule\midrule
        Q & 94.51 & 85.54 & 85.13 & 85.33 & 74.42\\
        Q+V & \textcolor{red}{\textbf{94.83}} & \textcolor{blue}{\underline{86.20}} & \textcolor{red}{\textbf{86.27}} & \textcolor{red}{\textbf{86.23}} & \textcolor{red}{\textbf{75.80}} \\
        Q+K+V+O & \textcolor{blue}{\underline{94.73}} & \textcolor{red}{\textbf{86.35}} & \textcolor{blue}{\underline{85.62}} & \textcolor{blue}{\underline{85.98}} & \textcolor{blue}{\underline{75.41}}\\
	\bottomrule
\end{tabular}}
\label{tab:rank_layer}
    } 
\end{tabular}
\label{tab:rank_ablation}
\end{table*}

\begin{table}[!t]
\centering
\caption{The comparison of UrbanSAM with advanced algorithms regarding total model parameters, fine-tuning parameters, and resource usage. Our results are presented in varying shades of gray, corresponding to the different parameter sizes of the three versions. The best and second-best results are highlighted in bold red and underlined blue, respectively.}
\setlength{\tabcolsep}{5.5pt}
\small
\begin{tabular}{cccccc}
\toprule[1.0pt]
\multirow{2}{*}{Methods} & \multirow{2}{*}{Backbone} & {Total} & {Learnable} & {IoU}  \\
\cmidrule(r){3-5}
  &  & {  (M)} & {  (M)} & {(\%)}   \\
\midrule\midrule
UNet & UNet64 & 17.26 & 17.26 & 71.93  \\
DeepLabv3+ & ResNet50 & 39.76 & 39.76 & 67.63 \\
HRNetv2 & HRNetv2-48 & 65.85 & 65.85 & 70.09\\
U2Net & U2Net-full & 44.00 & 44.00 & 72.05  \\
MAPNet & MAPNet & 23.69 & 23.69 & 71.23  \\
CBRNet & VGG16 & 22.69 & 22.69 & 72.27  \\
CGSANet & ResNet34 & 43.03 & 43.03 & 75.13 \\
HD-Net & HD-Net & 13.89 & 13.89 & 75.53  \\
Segformer & Segformer-B & 84.59 & 84.59 & 70.97  \\
Uformer & Uformer-B & 50.39 & 50.39 & 70.87  \\
BuildFormer & BuildFormer & 38.35 & 38.35 & 75.74  \\
\midrule
\multirow{3}{*}{SAM} & ViT-B & 90.49 & 90.49 & 70.18 \\
 & ViT-L & 308.01 & 308.01 & 71.30 \\
  & ViT-H & 635.64 & 635.64 & 71.33 \\
  \midrule
\multirow{3}{*}{HQ-SAM} & ViT-B & 91.56 & 1.07 & 71.03 \\
  & ViT-L & 309.34 & 1.33 & 72.14 \\
  & ViT-H & 637.23 & 1.59 & 71.92 \\
  \midrule
\multirow{2}{*}{RSBuilding} & ViT-B & 99.79 & 99.79 & 73.59 \\
 & ViT-L & 319.77 & 319.77 & 73.41 \\
\midrule 
\multirow{3}{*}{{UrbanSAM}} & \cellcolor{gray!15}ViT-B & \cellcolor{gray!15}100.91 & \cellcolor{gray!15}11.20 & \cellcolor{gray!15}75.80  \\
 & \cellcolor{gray!35} ViT-L & \cellcolor{gray!35} 326.09 & \cellcolor{gray!35}18.89 & \cellcolor{gray!35}\textcolor{blue}{\underline{\textbf{76.07}}}\\
 & \cellcolor{gray!60}ViT-H & \cellcolor{gray!60}663.49 & \cellcolor{gray!60}28.71 & \cellcolor{gray!60}\textcolor{red}{\textbf{76.63}} \\
\bottomrule[1.0pt]
\end{tabular}\label{tab:complexity}
\end{table}

\subsection{Ablation Analysis}
\subsubsection{Effectiveness of Learned U-Scaling Adapter}
To evaluate the effectiveness of the mask prompts learned by our invariance-based adapter, we simulated various prompt scenarios with different accuracy levels: mask prompts were generated by random erosion and dilation; point prompts were created from randomly distributed positive and negative cue points; and box prompts were generated through random offsets. The overlap rate with GT quantified the accuracy of the prompt generator. IoU measures the mask and box tips, and the point prompt accuracy is based on the matching ratio to the GT point.

Tab. \ref{tab:prompter} shows that UrbanSAM is not only comparable to the mask prompt with 70$\%$ accuracy and the 20 points prompt with $90\%$ accuracy, but even better than the box prompt with $100\%$ accuracy. Fig. \ref{fig:show_tab} presents the segmentation results of UrbanSAM for water, road, and building on a global dataset. This suggests that scale-invariance-based prompt strategies have significant generalization ability and adaptability to effectively respond to different ground object extraction tasks.

\subsubsection{Individual Components}
As shown in Tab. \ref{tab:lora}, our ablation experiments on the Massachusetts building dataset verified the applicability of the UrbanSAM module for large-scale urban fine mapping. The experimental results showed that the LoRA fine-tuning strategy significantly enhanced the zero-sample transmission capacity of SAM, increasing by 1.24 $\%$ compared with when LoRA was not used. Compared with the single-scale cue, the IoU score was increased by 0.83$\%$, indicating its advantage in fine-grained ground-extraction tasks. The cross-attention module achieves a 0.43$\%$ improvement in baseline performance by effectively integrating spatial details and global semantics. The hierarchical decoder structure is particularly prominent in the recovery of segmentation details, which further improves the accuracy of ground object extraction by nearly 2$\%$. These results show that the various modules of UrbanSAM play a key role in enhancing the model segmentation accuracy and detail fidelity.

To further evaluate the contribution of the multiscale operation and mask cross-attention module to the UrbanSAM model in the ground object information extraction, we performed class activation mapping (CAMs) visualization on the Massachusetts building dataset, and the results are shown in Fig. \ref{fig:abl_scale} and Fig. \ref{fig:abl_cross}. Compared with the single-scale operation and non-interaction-free approaches, multiscale UrbanSAM based on cue-and-interaction focuses more on key semantic regions and shows greater capture of small objects and boundary details, highlighting its advantages in fine object extraction tasks. This performance benefits from the larger receptive field and richer context features brought by the multiscale operation structure, which significantly improves the model's ability to identify ground objects at different scales. Moreover, mask cross-attention effectively enhances the model's performance in detail extraction and overall segmentation accuracy by establishing complementarity and correlation between modal feature sequences.

\subsubsection{LoRA utilization strategy}
Although existing research has explored LoRA fine-tuning strategies in the medical field, the approach largely remains empirical in remote sensing. Tab. \ref{tab:lora} presents the fine-tuning results for different encoder-decoder configurations. It is observed that applying LoRA to both the encoder and decoder outperforms strategies where LoRA is used only to the decoder or where all decoder parameters are fine-tuned, despite the latter involving a larger number of learnable parameters. This may be attributed to the differences between remote sensing object extraction and traditional computer vision segmentation tasks, where excessive parameter updates may overwrite the inherent segmentation priors of SAM, leading to catastrophic forgetting.

Tab. \ref{tab:rank_size} explores the impact of different rank sizes for the LoRA layers on UrbanSAM performance. The optimal performance is achieved when the rank is set to 4, with further increases resulting in performance degradation. Tab. \ref{tab:rank_layer} demonstrates that applying LoRA solely to the q and v projection layers yields the best performance, further highlighting that excessive parameter updates during domain adaptation can increase the complexity and difficulty of training.

\begin{figure}[!t]
\centering
\includegraphics[width=0.45\textwidth]{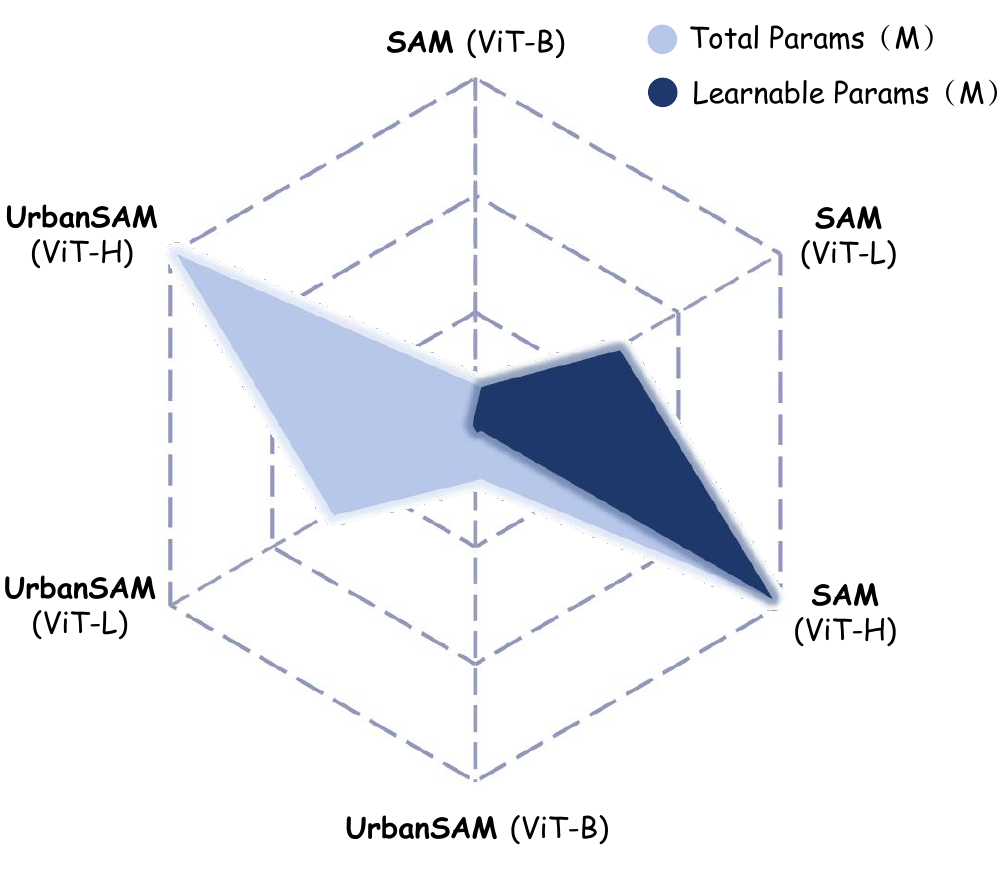}
\caption{Radar charts comparing SAM and UrbanSAM across three versions: base, large, and huge, with respect to parameter size. Light blue represents the total model parameters, while dark blue indicates the learnable (fine-tuned) parameters.}
\label{fig:complex}
\end{figure}

\subsubsection{Complex Analysis}
To further validate the effectiveness and demonstrate the advantages of UrbanSAM, we analyzed the network complexity on the Massachusetts Building dataset. Tab. \ref{tab:complexity} and Fig. \ref{fig:complex} show that the ViT-Base version of UrbanSAM achieves the highest segmentation accuracy (75.80$\%$) with the least learnable parameters (11.20M) in multiple cue-independent network frames including CNN, Transformer and SAM structural variants. Although HQ-SAM also has few learnable parameters, its performance relies on artificial cues, which often brings about additional parameter consumption and performance fluctuations. Notably, the ViT-Base and ViT-Large versions of UrbanSAM improved by 2.21$\%$ and 3.66$\%$ on the IoU using only 11.1$\%$ and 5.8$\%$ of the total parameters over the RSbuilding models trained with all parameters. Further, it shows that UrbanSAM significantly improves the feature extraction accuracy with only a few additional computational costs, demonstrating its significant advantages in urban fine-ground object mapping.

\section{Conclusion}
This paper introduces UrbanSAM, the first segmentation foundation model designed to autonomously address the fine-grained complexity and morphological heterogeneity of global urban scenes. Unlike existing SAM optimization strategies that often focus on fine-tuning decoders or utilizing mismatched pre-trained encoders, UrbanSAM incorporates a series of flexible and learnable Uscaling-adapters inspired by MRA theory. These adapters are seamlessly integrated into the trunk encoder to combine domain-specific priors and multi-resolution invariance with the general knowledge of pre-trained foundation models. This capability to capture and learn inherently invariant properties ensures strong applicability and robustness across globally diverse urban environments.

This seamless inheritance is achieved through hierarchical cross-attention and LoRA operations, which effectively align and fuse multiscale adapters with the trunk encoder, eliminating the reliance on manual interactive prompting and yielding learnable prompts. Extensive experimental results on a global-scale dataset demonstrate UrbanSAM's flexibility and superior segmentation performance in handling scale-varying urban objects, including buildings, roads, and water bodies.

Looking ahead, our research aims to achieve several key objectives: expanding the quantity and diversity of training data by incorporating a broader range of types, modalities, and temporal sequences. These enhancements will significantly improve the model's versatility and unlock its potential for many real-world applications.

\bibliographystyle{IEEEtran}
\bibliography{ref}

\begin{IEEEbiography}[{\includegraphics[width=1in,height=1.25in,clip,keepaspectratio]
{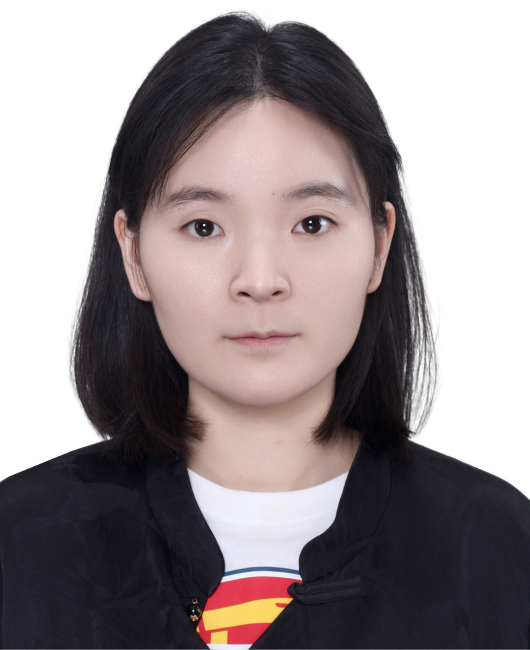}}]{Chenyu Li} received the B.S. and M.S. degrees from the School of Transportation, Southeast University, Nanjing, China, in 2018 and 2021, respectively. She is currently pursuing her Ph.D. degree in mathematics at Southeast University, Nanjing, China. She is also a joint Ph.D. student at the Aerospace Information Research Institute, Chinese Academy of Sciences, Beijing, China. Her research interests include interpretable artificial intelligence, big Earth data forecasting, foundation models, and hyperspectral imaging.
\end{IEEEbiography}

\vskip -2\baselineskip plus -1fil

\begin{IEEEbiography}[{\includegraphics[width=1in,height=1.25in,clip,keepaspectratio]{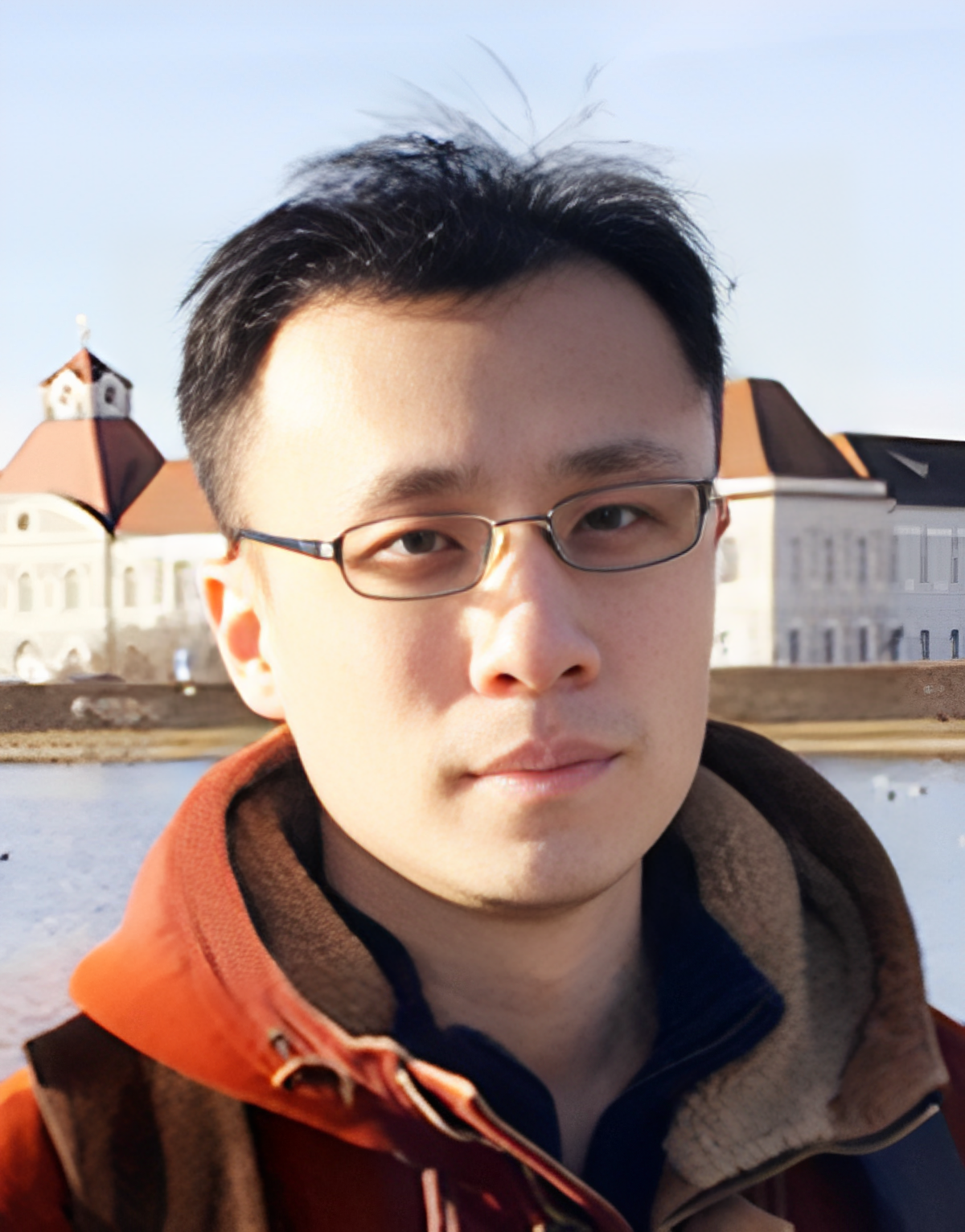}}]{Danfeng Hong}(IEEE Senior Member) received the Dr. -Ing degree (summa cum laude) from the Signal Processing in Earth Observation (SiPEO), Technical University of Munich (TUM), Munich, Germany, in 2019. Since 2022, he has been a Full Professor with the Aerospace Information Research Institute, Chinese Academy of Sciences. His research interests include Artificial Intelligence, Multimodal Big Data, Foundation Models, and Earth Observation. Dr. Hong is an Associate Editor for the IEEE Transactions on Image Processing (TIP) and the IEEE Transactions on Geoscience and Remote Sensing (TGRS). He is also an Editorial Board Member for Information Fusion and the ISPRS Journal of Photogrammetry and Remote Sensing. He has been recognized as a Highly Cited Researcher by Clarivate Analytics in 2022, 2023, and 2024.
\end{IEEEbiography}

\vskip -2\baselineskip plus -1fil

\begin{IEEEbiography}[{\includegraphics[width=1in,height=1.25in,clip,keepaspectratio]{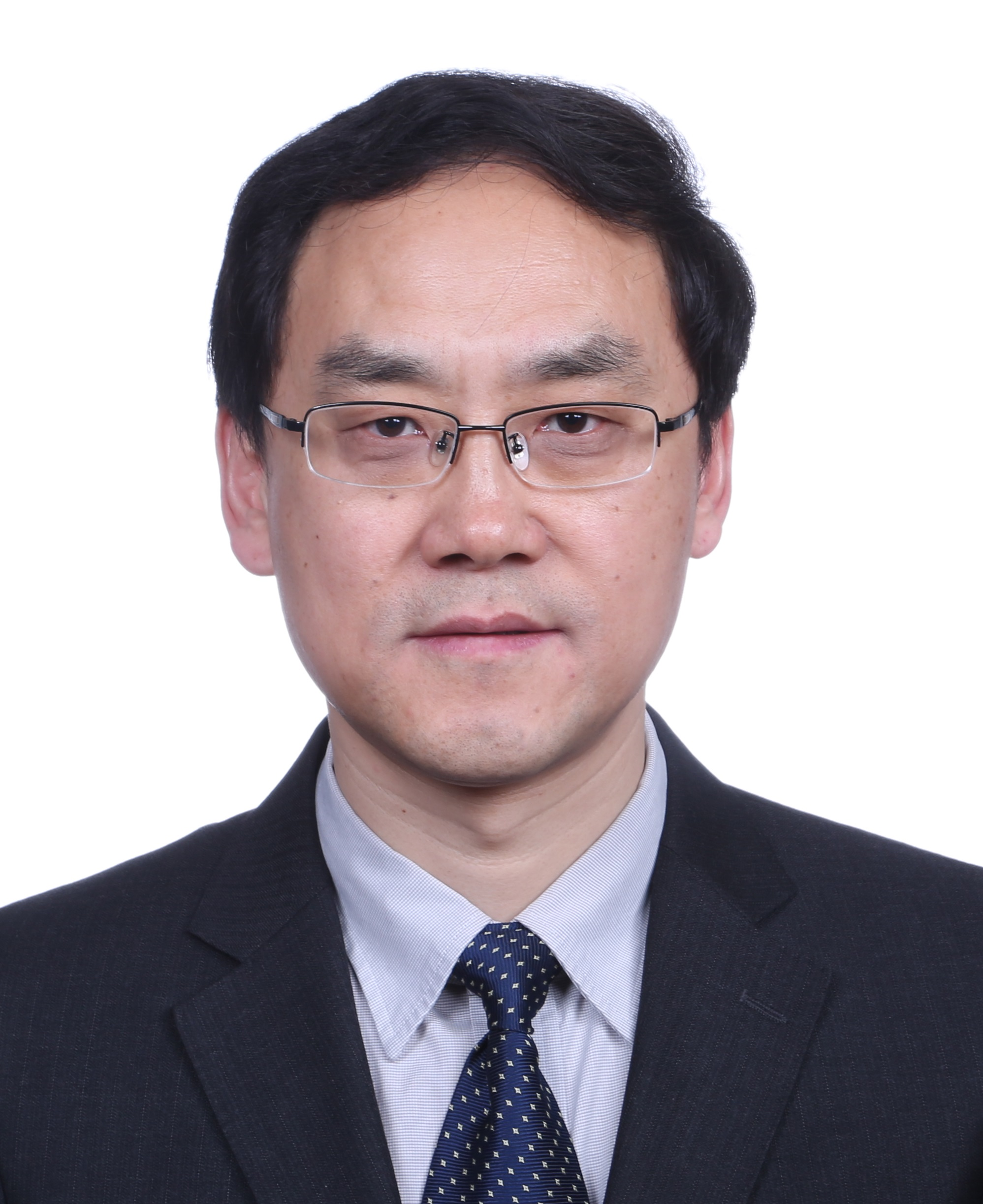}}]{Bing Zhang}(IEEE Fellow) received the B.S. degree in geography from Peking University, Beijing, China, in 1991, and the M.S. and Ph.D. degrees in remote sensing from the Institute of Remote Sensing Applications, Chinese Academy of Sciences, Beijing, in 1994 and 2003, respectively. He is currently a Full Professor and the Deputy Director of the Aerospace Information Research Institute, China Academy of Sciences. He has long been engaged in research on Hyperspectral remote sensing technology and applications. His creative achievements were rewarded with more than 10 important prizes, including the IEEE Geoscience and Remote Sensing Society (GRSS) Regional Leader Award, the National Science and Technology Advance Award of China, the Outstanding Scientific Achievement Award of the Chinese Academy of Sciences, etc. 
\end{IEEEbiography}

\vskip -2\baselineskip plus -1fil

\begin{IEEEbiography}[{\includegraphics[width=1in,height=1.25in,clip,keepaspectratio]{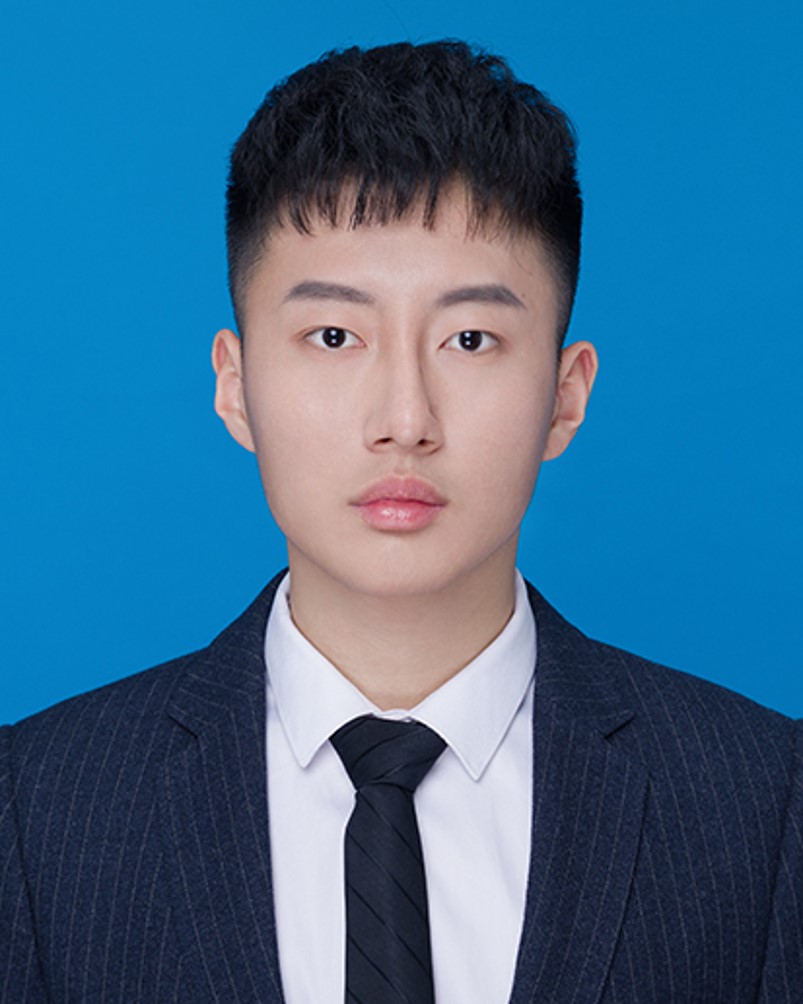}}]{Yuxuan Li} received the B.E. degree from the Institute of Geophysics and Geomatics, China University of Geosciences, Wuhan, China, in 2023. He is currently pursuing his Ph.D. degree at the Aerospace Information Research Institute, Chinese Academy of Sciences, Beijing, China. His research interests include deep learning and remote sensing image processing.
\end{IEEEbiography}

\vskip -2\baselineskip plus -1fil

\begin{IEEEbiography}[{\includegraphics[width=1in,height=1.25in,clip,keepaspectratio]{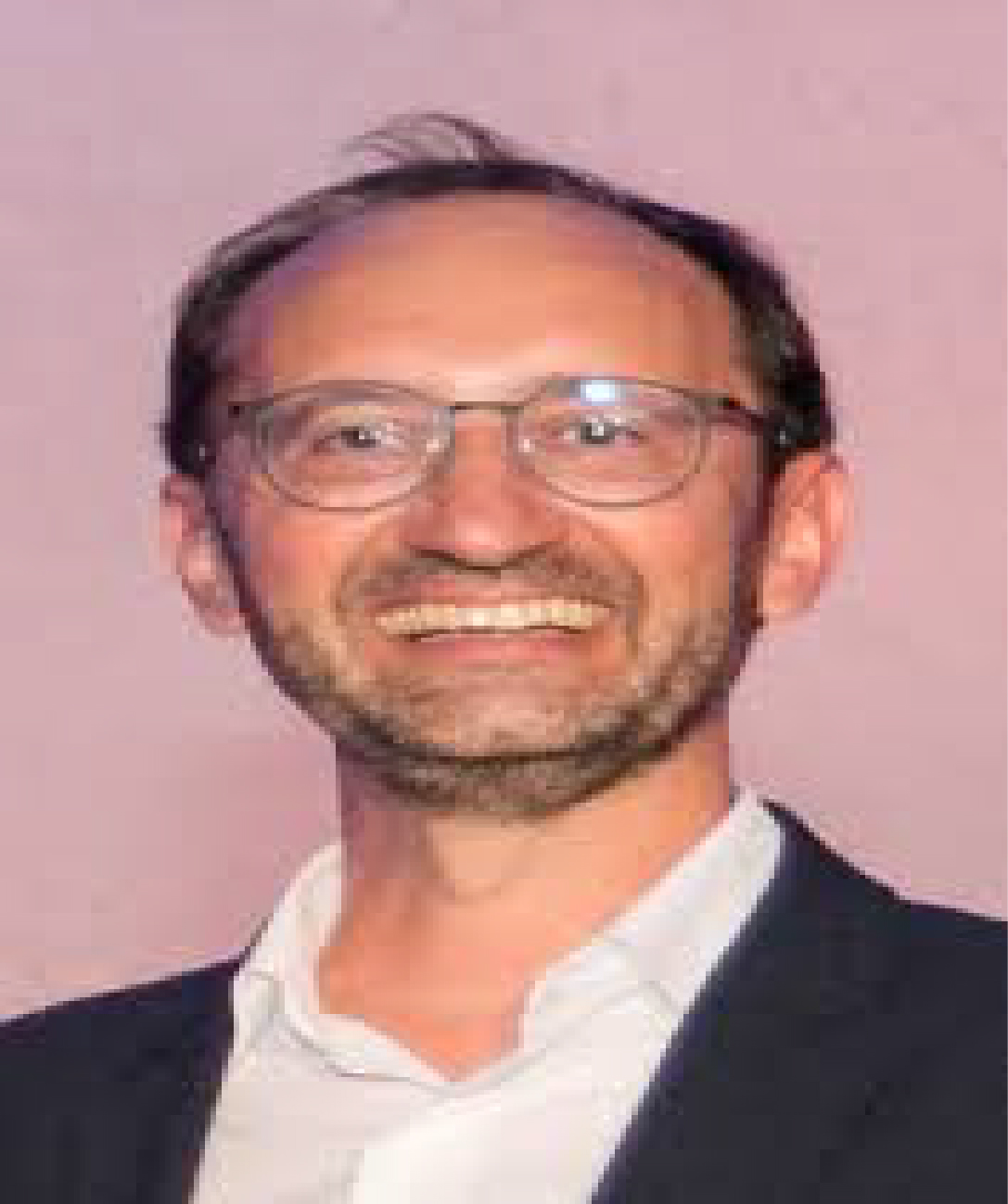}}]{Gustau~Camps-Valls} (IEEE Fellow) is a Full Professor of Electrical Engineering at Universitat de València, Spain. He specializes in AI and machine learning for environmental and climate challenges. He heads the Image and Signal Processing (ISP) group, focusing on AI for Earth and climate sciences. His research encompasses AI for predicting extreme events, enhancing Earth models, and understanding climate impacts on society, such as climate-induced migration. He is a Highly Cited Researcher (2011, 2021-2024) and has received notable recognitions, including the Google Classic Paper Award (2019). He has been the Program Chair for IEEE IGARSS 2018 and AISTATS 2022, an Associate Editor for five IEEE journals, and an IEEE Distinguished Lecturer (2017-2019). He is a Fellow of IEEE, ELLIS, EurASc, Academia Europaea, and AAIA. His leadership extends to advisory roles, including at ESA PhiLab and EUMETSAT, and he coordinates ELLIS's AI program for Earth and climate sciences. He received two ERC grants (consolidator and synergy) to advance AI for Earth and climate sciences.
\end{IEEEbiography}

\vskip -2\baselineskip plus -1fil

\begin{IEEEbiography}[{\includegraphics[width=1in,height=1.25in,clip,keepaspectratio]{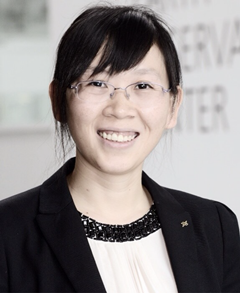}}]{Xiao Xiang Zhu} (IEEE Fellow) received the master (M.Sc.), doctor of engineering (Dr.-Ing.), and ``Habilitation'' degrees in signal processing from the Technical University of Munich (TUM), Munich, Germany, in 2008, 2011, and 2013, respectively. She is the Chair Professor for Data Science in Earth Observation with TUM and was the founding Head of the Department ``EO Data Science,'' the Remote Sensing Technology Institute, German Aerospace Center (DLR). Since May 2020, she has been the PI and Director of the international future AI lab ``AI4EO – Artificial Intelligence for Earth Observation: Reasoning, Uncertainties, Ethics and Beyond,'' Munich, Germany. Since October 2020, she has also been a Director of the Munich Data Science Institute (MDSI), TUM.  Her research interests include remote sensing and Earth observation, signal processing, machine learning, and data science. Dr. Zhu is a Fellow of the Academia Europaea (the Academy of Europe). She is an Associate Editor of IEEE Transactions on Geoscience and Remote Sensing, Pattern Recognition and was the Area Editor responsible for special issues of IEEE Signal Processing Magazine (2021–2023). She is a Fellow of AAIA and ELLIS.
\end{IEEEbiography}

\vskip -2\baselineskip plus -1fil

\begin{IEEEbiography}[{\includegraphics[width=1in,height=1.25in,clip,keepaspectratio]{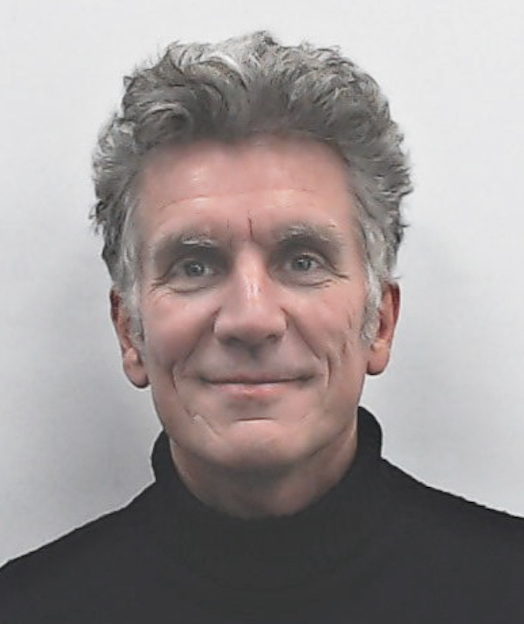}}]{Jocelyn Chanussot} (IEEE Fellow) received the M.Sc. degree in electrical engineering from the Grenoble Institute of Technology (Grenoble INP), Grenoble, France, in 1995, and the Ph.D. degree from the Université de Savoie, Annecy, France, in 1998. From 1999 to 2023, he was with Grenoble INP, where he was a Professor of signal and image processing. He is currently a Research Director with INRIA, Grenoble. His research interests include image analysis, hyperspectral remote sensing, data fusion, machine learning, and artificial intelligence.

Dr. Chanussot is the founding President of the IEEE Geoscience and Remote Sensing French chapter. He was the Vice-President of the IEEE Geoscience and Remote Sensing Society, in charge of meetings and symposia. He is an Associate Editor for the IEEE Transactions on Geoscience and Remote Sensing, the IEEE Transactions on Image Processing, and the Proceedings of the IEEE. He was the Editor-in-Chief of the IEEE Journal of Selected Topics in Applied Earth Observations and Remote Sensing (2011-2015). He is a Fellow of the IEEE, an ELLIS Fellow, a Fellow of AAIA, a member of the Institut Universitaire de France (2012-2017), and a Highly Cited Researcher (Clarivate Analytics/Thomson Reuters, since 2018).
\end{IEEEbiography}

\end{document}